\RequirePackage{fix-cm}
\RequirePackage{fixltx2e}
\documentclass[twoside,english,aip, jcp,floatfix]{revtex4-1}
\usepackage{mathpazo}

\usepackage[T1]{fontenc}
\usepackage[latin9]{inputenc}
\usepackage[a4paper]{geometry}
\usepackage{xcolor}
\usepackage{babel}
\usepackage{amsmath}
\usepackage{amssymb}
\usepackage{stmaryrd}
\usepackage{graphicx}
\usepackage[unicode=true,pdfusetitle,
 bookmarks=false,
 breaklinks=true,pdfborder={0 0 0},pdfborderstyle={},backref=false,colorlinks=true]
 {hyperref}
\hypersetup{
 citecolor = blue, linkcolor = blue,  urlcolor  = blue}
\usepackage{breakurl}

\makeatletter

\providecommand{\tabularnewline}{\\}

\usepackage{orcidlink}

\makeatother

\begin{document}

\title{{\Large{}Calibration in Machine Learning Uncertainty Quantification:
beyond consistency to target adaptivity}}

\author{Pascal PERNOT \orcidlink{0000-0001-8586-6222}}

\affiliation{Institut de Chimie Physique, UMR8000 CNRS,~\\
Université Paris-Saclay, 91405 Orsay, France}
\email{pascal.pernot@cnrs.fr}

\begin{abstract}
\noindent Reliable uncertainty quantification (UQ) in machine learning
(ML) regression tasks is becoming the focus of many studies in materials
and chemical science. It is now well understood that \emph{average}
calibration is insufficient, and most studies implement additional
methods testing the \emph{conditional} calibration with respect to
\emph{uncertainty}, i.e. \emph{consistency}. Consistency is assessed
mostly by so-called reliability diagrams. There exists however another
way beyond average calibration, which is\emph{ conditional} calibration
with respect to \emph{input features}, i.e. \emph{adaptivity}. In
practice, adaptivity is the main concern of the final users of a ML-UQ
method, seeking for the reliability of predictions and uncertainties
for any point in features space. This article aims to show that consistency
and adaptivity are complementary validation targets, and that a good
consistency does not imply a good adaptivity. An integrated validation
framework is proposed and illustrated on a representative example. 
\end{abstract}
\maketitle
\clearpage{}

\section{Introduction}

\noindent The quest for trust or confidence in the predictions of
data-based algorithms\citep{Vishwakarma2021,Gruich2023,Heid2023,Torrisi2023}
has led to a profusion of uncertainty quantification (UQ) methods
in machine learning (ML).\citep{Pearce2018,Musil2019,Hirschfeld2020,Tran2020,Abdar2021,Gawlikowski2021,Tynes2021,Zelikman2021,Hu2022,Varivoda2022,Battaglia2023,Busk2023_arXiv,He2023,Mohanty2023,Tohme2023}
However, not all of these UQ methods provide uncertainties that can
be relied upon,\citep{Liu2021,Pernot2022b} notably if, as in metrology,
one expects uncertainty to inform us on a range of plausible values
for a predicted property.\citep{GUM,Irikura2004}

In pre-ML computational chemistry, UQ metrics consisted essentially
in \emph{standard uncertainty}, i.e. the standard deviation of the
distribution of plausible values (a \emph{variance-based} metric),
or \emph{expanded uncertainty}, i.e. the half-range of a prediction
interval, typically at the 95\,\% level (an \emph{interval-based}
metric).\citep{Ruscic2004,Irikura2004} The advent of ML methods provided
UQ metrics beyond this standard setup, for instance distances in feature
or latent space\citep{Janet2019,Hullermeier2021,Hu2022} or the $\Delta$-metric,\citep{Korolev2022}
which have no direct statistical or probabilistic meaning. These metrics
might however be converted to variance-based metrics by \emph{post
hoc} recalibration methods such as temperature scaling\citep{Guo2017,Kuleshov2018,Levi2022}
and isotonic regression\citep{Busk2022}, or to interval-based metrics
by conformal inference.\citep{Vovk2012,Angelopoulos2021,Cauchois2021,Feldman2021}
Nevertheless, all UQ metrics need to be validated to ensure that they
are adapted to their intended use. In this study, I focus on the reliability
of variance-based UQ metrics for the prediction of properties at the
individual level.\citep{Reiher2022} 

The validation of UQ metrics is based on the concept of \emph{calibration}.
A handful of validation methods exist that explore more or less complementary
aspects of calibration. A trio of methods seems to have recently taken
the center stage: the \emph{reliability diagram}\citep{Guo2017,Levi2022},
the \emph{calibration curve\citep{Kuleshov2018}} and the \emph{confidence
curve}\citep{Ilg2018,Scalia2020}. They implement three different
approaches to calibration which are not necessarily independent, but
it is essential to realize that they do not cover the full spectrum
of calibration requirements. In particular, none of these methods
addresses the essential reliability of predicted uncertainties with
respect to the input features, sometimes called \emph{individual}
calibration\citep{Chung2020,Zhao2020}. 

\subsection{Scope and limitations of the study}

\noindent The aim of this article is to propose a complete validation
framework for variance-based UQ metrics, based on the concept of \emph{conditional}
calibration and its complementary aspects of \emph{consistency} (conditional
calibration with respect to uncertainty) and \emph{adaptivity} (conditional
calibration with respect to input features). 

\begin{figure}[t]
\begin{centering}
\includegraphics[bb=0bp 0bp 780bp 900bp,clip,width=0.9\textwidth]{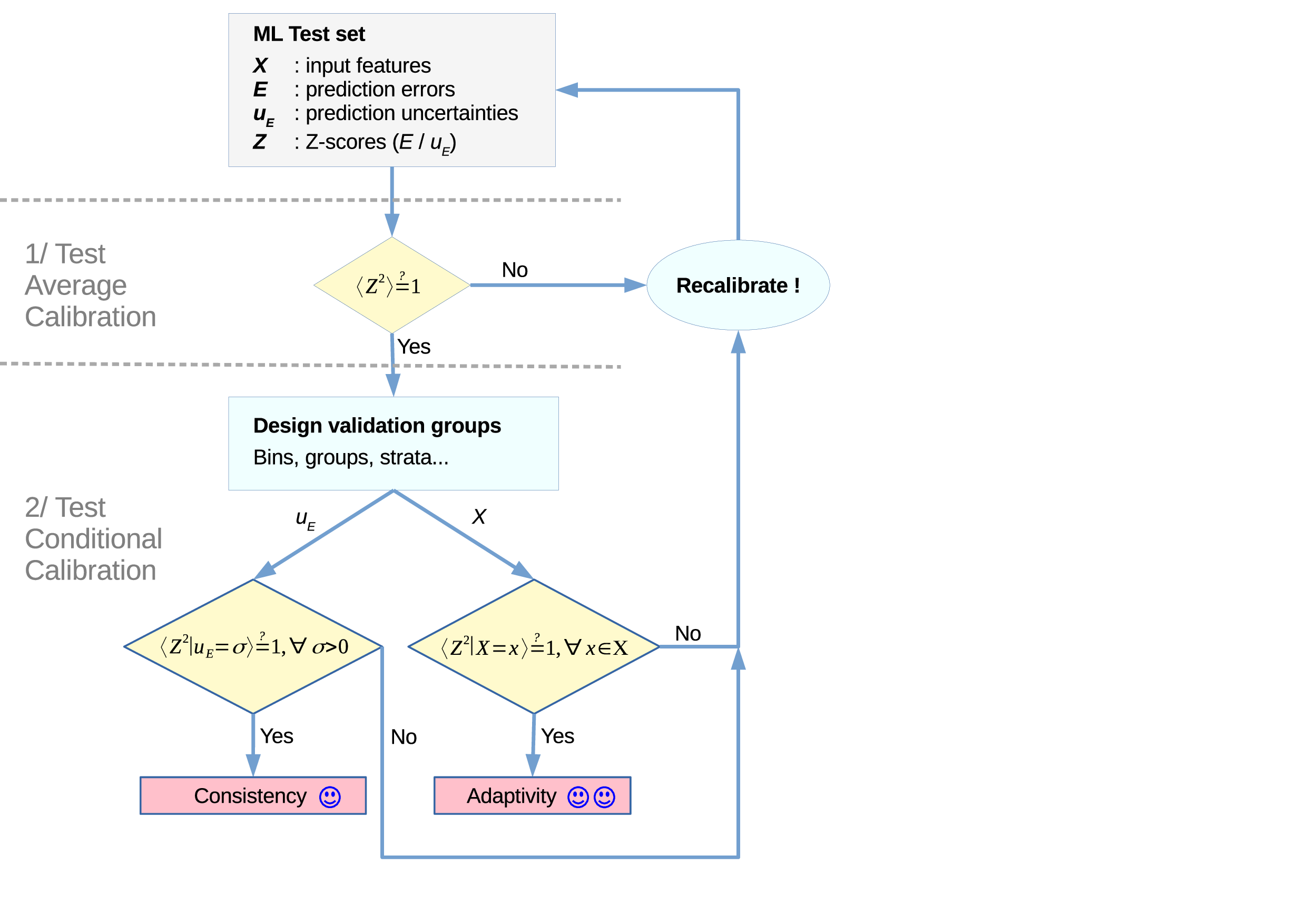}
\par\end{centering}
\caption{\label{fig:Flowchart}Flowchart of the \emph{z}-scores-based validation
framework.}

\end{figure}

It is well known that average calibration is not sufficient to establish
the reliability of ML-UQ predictions. This study goes one step further
and is designed to alert ML users that conditional calibration with
respect to uncertainty (consistency), as commonly tested by reliability
diagrams\emph{\citep{Guo2017,Laves2020,Levi2020,Scalia2020,Levi2022,Busk2022,Vazquez-Salazar2022},}
does not guarantee individual calibration. In order to approach individual
calibration, it is necessary to ensure conditional calibration with
respect to input features (adaptivity). As reliability diagrams are
not designed to deal with adaptivity, a more convenient statistical
framework dealing homogeneously with consistency and adaptivity is
proposed. The corresponding workflow is illustrated in Fig.\,\ref{fig:Flowchart}.

Note that for the sake of brevity, I present here only methods for
variance-based UQ metrics, but the approach can be directly transposed
to interval-based metrics\citep{Pernot2023_Arxiv}. Also, this study
does not offer advice nor recipes on how to achieve good conditional
calibration. 

\subsection{Structure of the article}

\noindent The next section (Sect.\,\ref{subsec:Validation-model-var})
introduces the notations and theoretical elements. The main notations
and acronyms are summarized in Table\,\ref{tab:Main-acronyms-used}.
The validation methods are presented in Sect.\,\ref{subsec:Consistency-testing}
and applied to a computational chemistry oriented example. The main
conclusions are presented in Sect.\,\ref{sec:Conclusion}. 
\begin{table}[t]
\noindent \begin{centering}
\textcolor{red}{}%
\begin{tabular}{lll}
\hline 
Notation &  & Definition\tabularnewline
\hline 
ACF &  & Auto-Correlation Function\tabularnewline
AGC &  & Adversarial Group Calibration\tabularnewline
CI &  & Confidence Interval (at the 95\% level)\tabularnewline
$E$ &  & Prediction error\tabularnewline
$f_{v,x}$ &  & Fraction of valid intervals for conditioning variable $x$\tabularnewline
MSE &  & Mean Squared Error: $<E^{2}>$\tabularnewline
(L)RCE &  & (Local) Relative Calibration Error: $\mathrm{(RMV-RMSE)/RMV}$\tabularnewline
(L)ZM &  & (Local) Z-scores Mean: $<Z>$\tabularnewline
(L)ZMS &  & (Local) Z-scores Mean Squared: $<Z^{2}>$\tabularnewline
RMSE &  & Root Mean Squared Error: $\sqrt{\mathrm{RMSE}}$\tabularnewline
RMV &  & Root Mean Variance: $\sqrt{<u_{E}^{2}>}$\tabularnewline
$u_{E}$ &  & Prediction uncertainty\tabularnewline
$X,\,X_{i}$ &  & Input feature\tabularnewline
$Z$ &  & \emph{Z}-score: $E/u_{E}$\tabularnewline
\hline 
\end{tabular}
\par\end{centering}
\caption{\label{tab:Main-acronyms-used}Main acronyms and notations used in
this study.}
\end{table}

\section{Validation of variance-based UQ metrics\label{subsec:Validation-model-var}}

\noindent The validation of variance-based UQ metrics requires\textcolor{orange}{{}
}at minimum a set of predicted values $V=\{V_{i}\}_{i=1}^{M}$, the
corresponding uncertainties $u_{V}=\{u{}_{V}\}_{i=1}^{M}$, and reference
data to compare with $R=\{R_{i}\}_{i=1}^{M}$ (with their uncertainties
$u_{R}=\{u_{R_{i}}\}_{i=1}^{M}$, when relevant). From these, one
estimates \emph{prediction errors} $E=R-V$ and \emph{prediction uncertainties}
$u_{E}=(u_{R}^{2}+u_{V}^{2})^{1/2}$. These data enable to test average
calibration (Sect.\,\ref{par:Average-calibration.}) and \emph{consistency}
as conditional calibration with respect to uncertainty (Sect.\,\ref{subsec:Individual,-conditional-and}).
An additional set of \emph{input features} or adequate proxies $X_{j}=\{X_{j,i}\}_{i=1}^{M}$
is required for a full validation setup including conditional calibration
with respect to inputs or \emph{adaptivity} (Sect.\,\ref{subsec:Individual,-conditional-and}).

\subsection{Average calibration\label{par:Average-calibration.}}

\noindent The validation of prediction uncertainty $u_{E}$ can be
based on the requirement that it correctly quantifies the \emph{dispersion}
of prediction errors $E$.\citep{Pernot2022a,Pernot2022b} Following
the metrological definition of uncertainty, this is valid for so-called
adequate models, i.e. models with negligible systematic or model errors.
For models with non-negligible inadequacy levels, as can be expected,
for instance, from ML methods\textcolor{red}{{} }below the interpolation
threshold\citep{Dar2021}, prediction uncertainty is often designed
to cover also for model errors.\citep{Pernot2017} In such cases,
validation should account for both bias and dispersion components
of the errors. I will not discuss here the problems of reporting such
uncertainties for actionable predictions or risk assessment, but will
mostly focus on a self-consistent setup for the validation of UQ calibration
where one seeks a reliable estimation of the amplitude of errors.

In such conditions, the basis for validation is to require that the
mean squared error (MSE) is close to the \emph{mean variance} over
the validation dataset\citep{Pernot2017} 
\begin{equation}
<E^{2}>\simeq<u_{E}^{2}>
\end{equation}
However, this formula ignores the essential one-to-one pairing of
errors and uncertainties, and a more stringent approach is based on
\emph{z}-scores ($Z=E/u_{E}$), using the condition
\begin{align}
<Z^{2}> & \simeq1\label{eq:Z2cal}
\end{align}
which is related to the Birge ratio\citep{Birge1932} for the validation
of the residuals of least-squares fit\citep{Pernot2022b}\textbf{.\vspace*{\bigskipamount}
Remarks.}
\begin{itemize}
\item If $E$ and $u_{E}$ are obtained as the means and standard deviations
of small ensembles of predictions (e.g. with less than 30 elements)
these formulas have to be adapted, and hypotheses need then to be
made on the error distributions for these small ensembles\citep{Pernot2022b}.
For a normal generative distribution of errors, the distribution of
the mean of $n$ values (ensemble size) is a Student's-\emph{t} distribution
with $\nu=n-1$ degrees of freedom, and one should have $<Z^{2}>-<Z>^{2}\simeq\nu/(\nu-2)$.\citep{Pernot2022b} 
\item Unbiasedness is not an essential part of calibration, but it is a
highly desirable property for predictions and \emph{z}-scores and
will be systematically considered as a test of prediction quality,
i.e.
\begin{equation}
<Z>\simeq0
\end{equation}
\item For unbiased \emph{z}-scores, one recovers the variance test proposed
previously by Pernot\citep{Pernot2022b} 
\begin{equation}
Var(Z)\simeq1\label{eq:ZV}
\end{equation}
which can be superseded by Eq.\,\ref{eq:Z2cal}.
\end{itemize}
The satisfaction of Eq.\,\ref{eq:Z2cal} validates \emph{average}
\emph{calibration}, which is a necessary requirement, but does not
guarantee the reliability of individual uncertainties, as average
calibration can be satisfied by a compensation of under- and over-estimated
values of $u_{E}$. The next section presents an approach to more
local validation techniques.\textbf{ \bigskip{}
}

\noindent \textbf{$\vartriangleright$ Example}. In a recent article,
Busk \emph{et al.}\citep{Busk2022} extended a message passing neural
network in order to predict properties of molecules and materials
with a calibrated probabilistic predictive distribution. A \emph{post
hoc} isotonic regression on data unseen during the training was used
to ensure calibration and was applied to materials science datasets.
I consider here the QM9 test dataset, which consists of $\text{M=13\thinspace885}$
atomization energies ($V,\thinspace u_{V}$), reference values ($R$),
and molecular formulas. 

Average calibration for this dataset is satisfactory, with $<Z^{2}>=0.96(2)\simeq1$,
and \emph{z}-scores are unbiased in average, with $<Z>=0.0082(83)\simeq0$.\hfill{}$\boxempty$

\subsection{Individual, conditional and local calibration\label{subsec:Individual,-conditional-and}}

\noindent The best calibration one could ideally achieve is \emph{individual}
calibration, a condition where one is confident that uncertainty is
correctly calibrated for any individual prediction. The formulation\emph{
}of individual calibration\emph{ }for probabilistic forecasters by
Chung\emph{ et al.}\citep{Chung2020}, led them to formalize it as
\emph{conditional calibration} \emph{in input features space.} In
practice (i.e. for finite size datasets), individual calibration has
been shown to be unreachable\citep{Zhao2020}, and an alternative
is to consider a discretized form as \emph{local }or\emph{ group}
calibration\citep{Luo2021}. This is reflected in the practical estimation
of conditional statistics by data binning or grouping\textcolor{orange}{\emph{\citep{Chung2020}}}.
In a similar spirit, \emph{conditional coverage} with respect to input
features was proposed by Vovk\citep{Vovk2012} to assess the \emph{adaptivity}\citep{Angelopoulos2021}
of conformal predictors.

For variance-based UQ metrics, Levi\emph{ et al.\citep{Levi2020,Levi2022}}
proposed\emph{ }an approach\emph{ }based on \emph{conditional calibration
in uncertainty space}, namely
\begin{equation}
<E^{2}|u_{E}=\sigma>\simeq\sigma^{2},\,\forall\sigma>0\label{eq:varEvalCond-1}
\end{equation}
which is the basis of the popular \emph{reliability diagrams\citep{Guo2017}}
or \emph{RMSE vs RMV} plots, also called \emph{calibration diagrams}\citep{Laves2020},
\emph{error-based calibration plots}\citep{Scalia2020}, \emph{RvE}
plots\citep{Palmer2022}, or \emph{RMSE vs. RMV} curves\citep{Vazquez-Salazar2022}.
Levi \emph{et al.} claim that, assuming that each uncertainty value
occurs only once in the dataset, their method ``\emph{captures the
desired meaning of calibration, i.e., for each individual example,
one can correctly predict the expected mistake}''.\textcolor{orange}{{}
}In practice, the unicity assumption faces two major difficulties:
(1) some datasets are \emph{stratified}, with several occurrences
of the same uncertainty value\citep{Pernot2023b_arXiv}, and (2) the
practical implementation of conditional calibration requires to group
data to estimate the mean squared error (MSE), breaking the one-to-one
correspondence between the tested uncertainties and errors, as mentioned
above for average calibration.\textcolor{orange}{{} }

In consequence, conditional calibration based on Eq.\,\ref{eq:varEvalCond-1}
is not sufficient to validate calibration at the individual level.
To go further, one should consider other conditioning variables besides
$u_{E}$, notably input features or variables of interest for the
end-user of a ML model, as proposed for probabilistic forecasters\textcolor{orange}{\emph{\citep{Chung2020}}}
and conformal predictors\citep{Vovk2012,Angelopoulos2021}. 

Building on the the works of Levi \emph{et al}.\emph{\citep{Levi2020},}
Pernot\citep{Pernot2022b} and Angelopoulos \emph{et al.}\citep{Angelopoulos2021}
about conditional calibration, I propose here to distinguish two calibration
targets (besides average calibration), namely \emph{consistency} as
the conditional calibration with respect to prediction uncertainty,
and \emph{adaptivity} as the conditional calibration with respect
to input features:
\begin{itemize}
\item \textbf{Consistency} is a special case of conditional calibration,
in the sense that it involves only $E$ and $u_{E}$. Using the \emph{z}-scores
statistics introduced for average calibration, one can define consistency
by the following equation
\begin{align}
<Z^{2}|u_{E}=\sigma> & \simeq1,\,\forall\sigma>0\label{eq:LZMS-cons}
\end{align}
Consistency is related to the metrological consistency of measurements\citep{Kacker2010}. 
\item \textbf{Adaptivity} is also conveniently formulated with \emph{z}-scores
as
\begin{align}
<Z^{2}|X=x> & \simeq1,\,\forall x\in\mathcal{X}\label{eq:LZMS-adapt}
\end{align}
where $\mathcal{X}$ is the ensemble of values accessible to $X$.
Adaptivity involves more information than consistency ($X$, $E$,
and $u_{E}$). 
\end{itemize}
Unless there is a monotonous transformation between $u_{E}$ and $X$,
consistency and adaptivity are distinct calibration targets, and a
good consistency does not augur of a good adaptivity and vice-versa,
so that both should be assessed. Note that \emph{tightness}, as introduced
earlier by Pernot\citep{Pernot2022b}, covers both consistency and
adaptivity. 

Average calibration is a necessary condition to reach consistency
or adaptivity. In fact, consistency/adaptivity expressed as conditional
calibration should imply average calibration, but the splitting of
the data into subsets makes that the power of individual consistency/adaptivity
tests is smaller than for the full validation set. It is therefore
better to test average calibration separately, notably for small validation
datasets. 

Most methods used to this day for the validation of variance-based
UQ metrics in chemical/materials sciences ML studies involve only
$E$ and $u_{E}$ (reliability diagrams, calibration curves, confidence
curves...)\citep{Scalia2020}. Adaptivity can thus be considered as
a blind spot in UQ validation, despite its necessity to achieve reliable
UQ at the molecule-specific level advocated by Reiher\citep{Reiher2022}. 

\section{Validation methods\label{subsec:Consistency-testing}}

\noindent This section presents \emph{z}-scores-based methods to assess
and validate consistency and adaptivity.\textcolor{red}{{} }An alternative
formulation, based on relative calibration errors, is also proposed
in Sect.\,\ref{subsec:LRCE}.

\subsection{Homoscedasticity plots of \emph{z}-scores}

\noindent A simple way to estimate consistency is to plot the \emph{z}-scores
$Z$ as a function of $u_{E}$ \citep{Pernot2022b}. The dispersion
of $Z$ should be homogeneous along $u_{E}$ (homoscedasticity) and,
ideally, symmetric around $Z=0$ (unbiasedness). In areas where the
\emph{z}-scores are biased, if any, one should observe a larger dispersion.
This might not be easy to appreciate visually, and \emph{running statistics
}can be superimposed to the data cloud such as the \emph{mean}, to
be compared with $Z=0$, and \emph{mean squares} to be compared with
the $Z=1$ line. 

This plot is easily extended to any variable $X$ other than $u_{E}$
and can be directly applied to the visual appreciation of adaptivity.
Note that in the present context, the $Z$ vs $u_{E}$ plot is preferable
to the $E$ vs $u_{E}$ plot used in other studies \citep{Janet2019,Pernot2022b,Hu2022},
as it offers a consistent representation for both consistency and
adaptivity estimation. 

For cases where consistency/adaptivity cannot be frankly rejected
on the basis of the shape or scale of this data cloud, it is necessary
to perform more quantitative tests as presented below. One should
not conclude on good consistency/adaptivity based solely on this kind
of plot.

\noindent \bigskip{}

\noindent \textbf{$\vartriangleright$ Example, continued.} The molecular
mass ($X_{1}$; in Dalton (Da)) and fraction of heteroatoms ($X_{2}$;
unitless) are generated from the molecular formulas of the QM9 dataset,
and used as proxies for input features. They are practically uncorrelated
between themselves and weakly correlated with $|E|$ and $u_{E}$
(Table\,\ref{tab:Rank-correlation-coefficients}). The dataset can
thus be tested for consistency and adaptivity. 
\begin{table}[t]
\noindent \centering{}%
\begin{tabular}{c|ccc}
 & $X_{2}$ & $|E|$ & $u_{E}$\tabularnewline
\hline 
$X_{1}$ & 0.12 & 0.10 & 0.03\tabularnewline
$X_{2}$ & - & 0.12  & 0.30\tabularnewline
$|E|$ & - & - & 0.32\tabularnewline
\end{tabular}\caption{\label{tab:Rank-correlation-coefficients}Rank correlation coefficients
for the\textcolor{violet}{{} }$\left\{ X_{1},X_{2},|E|,u_{E}\right\} $
set. }
\end{table}

\begin{figure*}[t]
\noindent \begin{centering}
\includegraphics[bb=0bp 0bp 1200bp 1100bp,clip,width=0.33\textwidth]{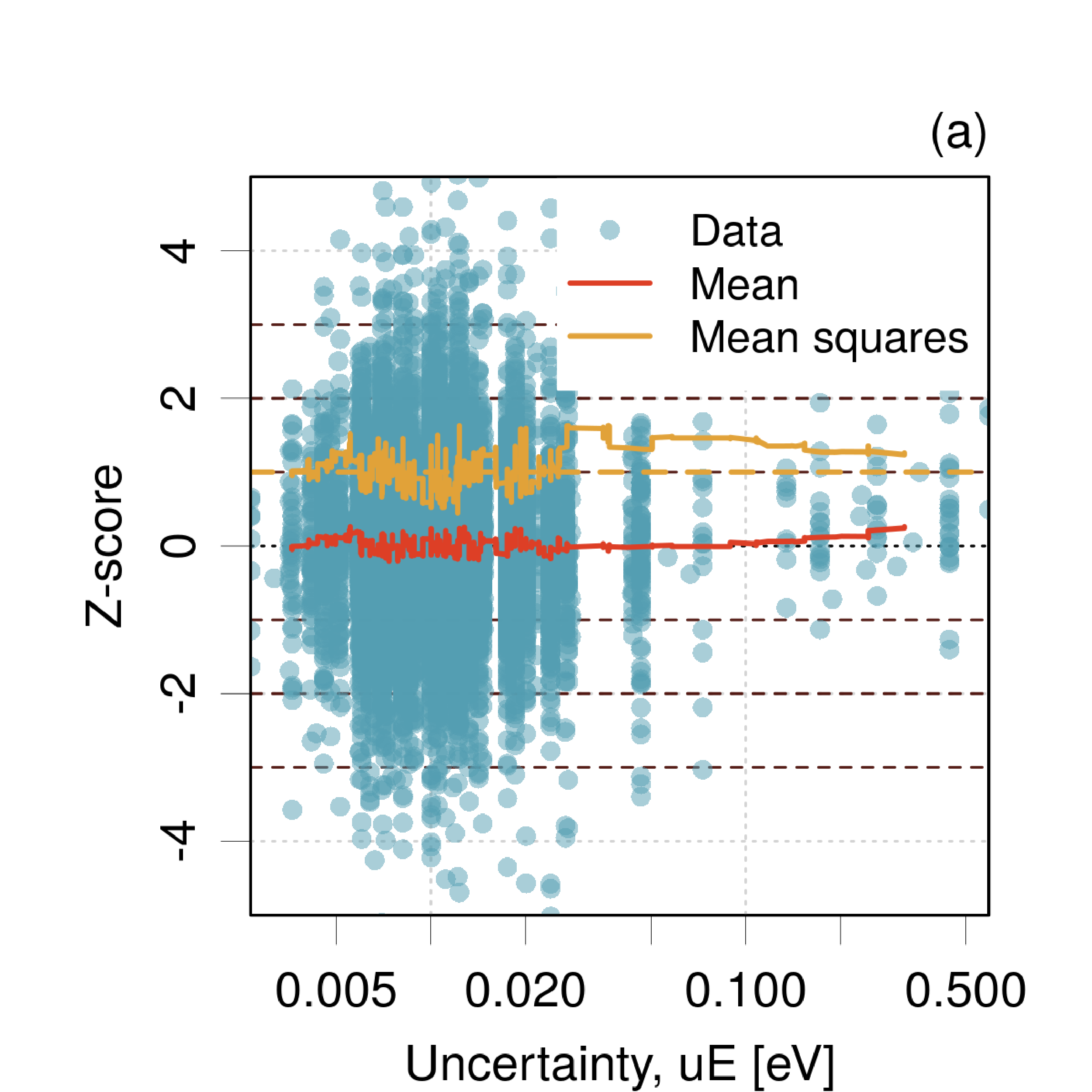}\includegraphics[bb=0bp 0bp 1200bp 1100bp,clip,width=0.33\textwidth]{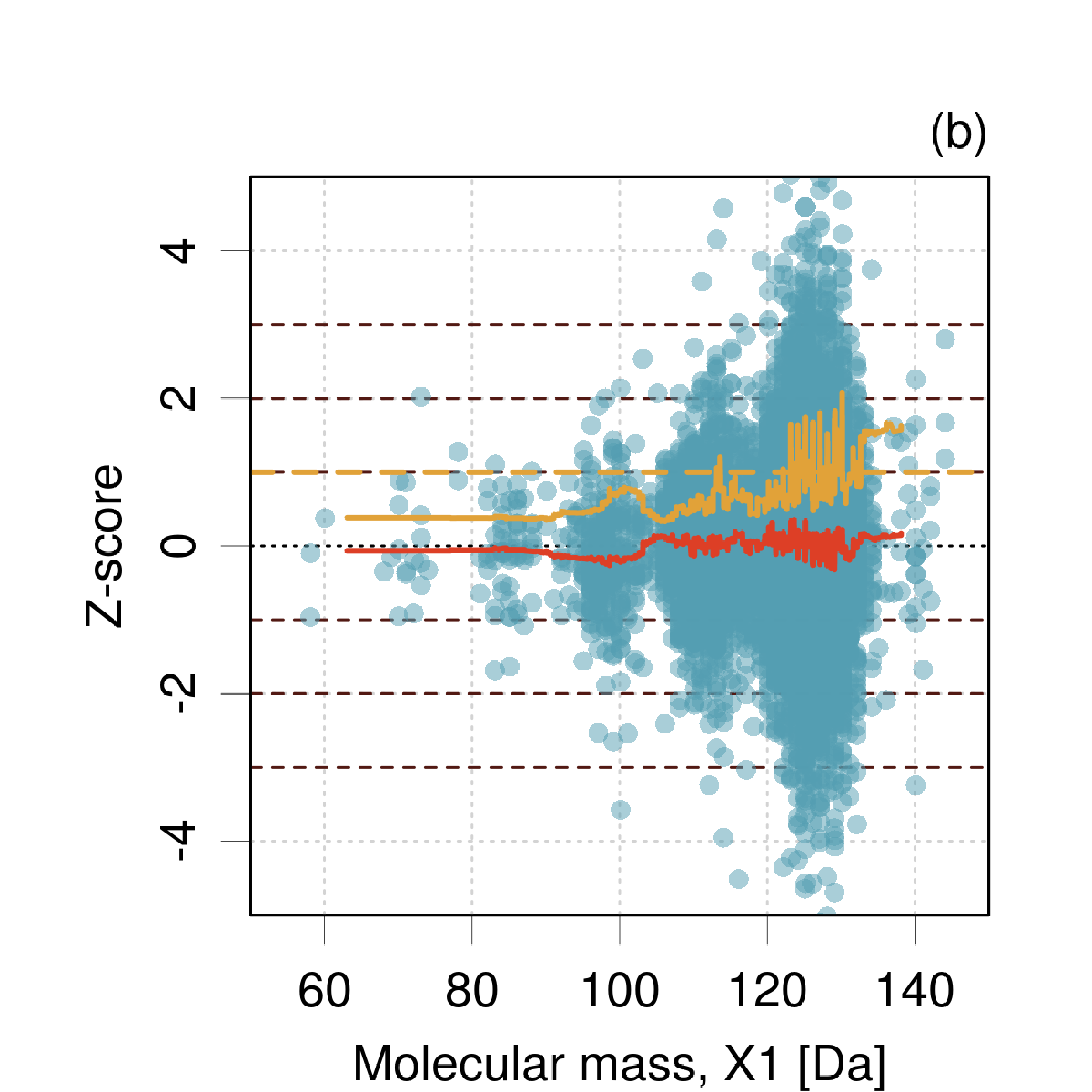}\includegraphics[bb=0bp 0bp 1200bp 1100bp,clip,width=0.33\textwidth]{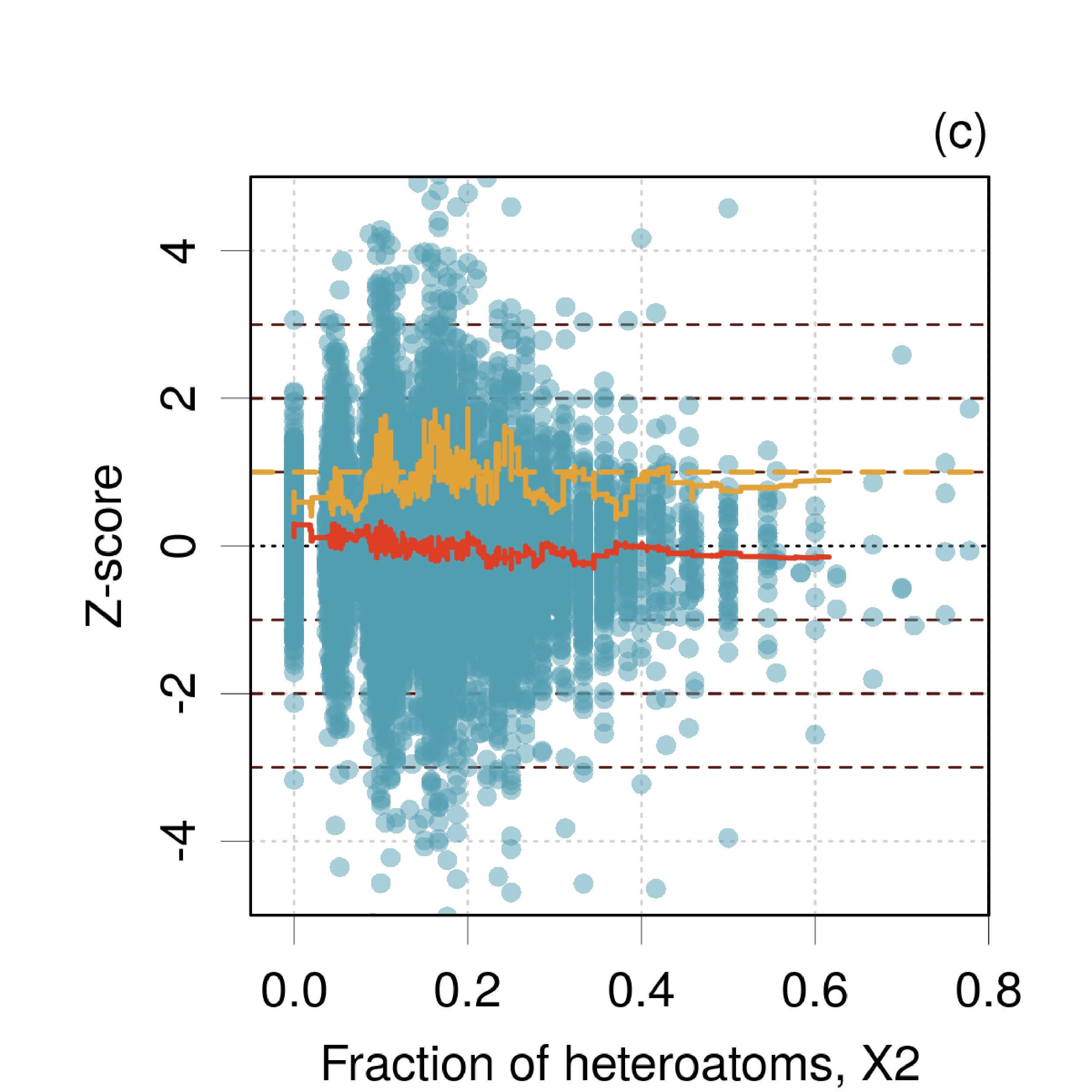}
\par\end{centering}
\caption{\label{fig:Plot-of-errors}QM9 dataset: \emph{z}-scores vs. uncertainty
(a), molecular mass (b) and the fraction of heteroatoms (c). Running
statistics (mean ($<Z>$) in red and mean squares ($<Z^{2}>$) in
orange) are estimated for a sliding window of size $M/100$.}
\end{figure*}

The homoscedasticity of \emph{z}-scores for the QM9 dataset is estimated
against $u_{E},$ $X_{1}$ and $X_{2}$ (Fig.\,\ref{fig:Plot-of-errors}).
One sees in Fig.\,\ref{fig:Plot-of-errors}(a) that the data points
are fairly symmetrically dispersed (mean; red line) and that the running
mean squares (orange line) follows rather closely the $Z=1$ line,
up to $uE\simeq0.02$\,eV, after which it lies at higher values.
However, this concerns a small population (980 points) and the problem
could be due to the data sparsity in this uncertainty range. 

The ``$Z$ vs $X_{1}$'' plot in Fig.\,\ref{fig:Plot-of-errors}(b)
enables to check if calibration is homogeneous in molecular mass space.
The running mean does not deviate notably from 0 (except around $X_{1}\simeq100$\,Da,
with a correlated increase in $<Z^{2}>$). The shape of the running
mean squares line, erring towards small $<Z^{2}>$ values, indicates
that uncertainties are probably overestimated for masses smaller than
the main mass cluster (around 125-130\,Da) evolving to a slight underestimation
above this peak. This trend hints at a lack of adaptivity. A similar
plot is shown for $X_{2}$, the fraction of heteroatoms {[}Fig.\,\ref{fig:Plot-of-errors}(c){]}
where the running mean presents a weak but systematic trend from positive
to negative values. Besides, the \emph{z}-scores are under-dispersed
for molecules with low heteroatoms fractions (below 0.1), after which
the running mean squares line presents notable oscillations around
the $Z=1$ reference line and seem to stabilize above $X_{2}=0.4$,
where the data are sparse.

In this dataset, stratification of the conditioning variables is notable.
For instance, the set of uncertainties $u_{E}$ contains only 138
distinct numerical values, a fact which can be attributed to recalibration
by a step-wise isotonic regression function. But stratification might
also occur independently of any algorithm: $X_{1}$ contains 398 unique
values, and $X_{2}$ is strongly stratified, with only 76 values.
Stratification should be taken into account when binning these variables
(see Sect. \ref{subsec:Binning/grouping-strategies}).

From these three plots, one gets the impression that calibration is
rather good at the core of the dataset (where the density of data
is highest), but more problematic in the margins. A more quantitative
analysis of these features is desirable, but one might already conclude
that adaptivity is not reached.\hfill{}{\scriptsize{} $\boldsymbol{\square}$} 

\subsection{Local calibration\label{subsec:Local-calibration-error}}

\noindent Conditional calibration in uncertainty space as formulated
in Eq.\,\ref{eq:varEvalCond-1} is often tested in the literature
by reliability diagrams based on groups defined as uncertainty bins\citep{Levi2022}.
This representation does not adapt conveniently to other grouping
schemes. In contrast, the approach based on \emph{z}-scores (Eqns.\,\ref{eq:LZMS-cons},\,\ref{eq:LZMS-adapt}),
besides its interest evoked for average calibration, offers a uniform
treatment for all conditioning variables and is used preferentially
in this study. For readers more familiar with the use of calibration
errors, an alternative formulation based on Local Relative Calibration
Error is proposed in Sect.\,\ref{subsec:LRCE}.

\subsubsection{Local Z-Mean and Z-Mean-Squares analysis\label{subsec:Local-Z-Variance-analysis}}

\noindent Testing for consistency is based on a binning of the data
according to increasing uncertainties. A Local Z Variance (LZV) analysis
was introduced by Pernot\citep{Pernot2022a} as a method to test local
calibration: for each bin, one estimates $\mathrm{Var}(Z)$ and compares
it to 1. In the present framework, the LZV analysis is adapted to
account for the possibility of accepting significant deviations of
local $<Z>$ values from 0, and one will be using the Local Z Mean
Squares (LZMS) statistic, based on Eq.\,\ref{eq:LZMS-cons}. A Local
Z Mean statistic can also be used to check the local unbiasedness
of \emph{z}-scores.

Assessment of a LZMS analysis is based on two criteria, the deviation
of the LZMS values from 1 and the homogeneity of their distribution
along the conditioning variable.

The maximal admissible deviations depend on the bin size and error
distribution. For instance, one should expect larger deviations from
errors and uncertainties obtained as statistical summaries of small
ensembles than from errors and uncertainties describing a normal distribution.
Unless the error model is well known and controlled, which might not
be the case for post-hoc calibration methods, it is impossible to
define a threshold to LZMS values for validation purpose. It is therefore
necessary to estimate confidence intervals (CIs) on the LZMS values
to test their consistency with the target value. 

For $<Z>$, the standard formula based on the quantiles of the Student's-\emph{t}
distribution provides intervals with satisfactory coverage, even for
small samples and non-normal distributions with finite variance. The
case of $<Z^{2}>$\textcolor{orange}{{} }is more difficult, as standard
formulas fare poorly when one deviates from the standard normality
of the \emph{z}-scores. The same problem was observed for $Var(Z)$,\citep{Pernot2022a}
and a convergence and power study concluded that the most reliable
approach was to use bootstrapping\citep{Efron1979,Efron1991} with
samples of at least 100 points. In such conditions, the effective
coverage of 95\,\% CIs reaches at least 90\,\% for $Var(Z)$ and
$<Z^{2}>$. To achieve a 95\% coverage with a Student's distribution
of \emph{z}-scores, 1000 points per bin are required, which might
limit the resolution of the local analysis. When in doubt, the LZMS
analysis can be performed with several bin sizes to assess its reliability.
When reliable CIs are obtained, the proportion of valid intervals,
i.e. those covering the target value, can be used as a validation
metric (Sect.\,\ref{subsec:Validation-metrics}).

The homogeneity of the distribution of the LZMS values along the conditioning
variable can often be appreciated visually (any cluster showing systematic
deviation from the target represents a local calibration problem).
However, the graph might sometimes be crowded, and the auto-correlation
function (ACF) of the LZMS statistics might help to detect the presence
of unsuitable serial correlations. 

\subsubsection{Validation metrics\label{subsec:Validation-metrics}}

\noindent Calibration metrics\citep{Maupin2018,Vishwakarma2021} are
widely used in the ML-UQ literature: for instance, metrics have been
designed for calibration curves\citep{Kuleshov2018,Tran2020}, reliability
diagrams\citep{Levi2020}, and confidence curves\citep{Scalia2020}.
These metrics are generally used to compare and rank UQ methods, but
they do not provide a validation setup accounting for the statistical
fluctuations due to finite-sized datasets or bins numbers. It has
been shown recently\citep{Pernot2023a_arXiv}, that the \emph{expected
normalized calibration error} (ENCE)\citep{Levi2020,Scalia2020,Vazquez-Salazar2022}
cannot be used directly as a validation metric. The calculation of
reference values for those metrics is an option introduced recently\citep{Pernot2022c,Rasmussen2023},
but being based on a probabilistic model, it requires the choice of
a probability distribution for the errors which might complicate the
diagnostic. 

Here, let us take advantage of the availability of confidence intervals
on the local statistics of the LZM and LZMS methods as the basis for
a validation metric. For a perfectly calibrated dataset, the fraction
of binned statistics with a CI containing the target value should
be close to the coverage probability of the CIs. Namely, about 95\,\%
of the binned $<Z^{2}>$ values should have 95\,\% CIs containing
the target value. Let us denote this fraction of validated intervals
by $f_{v,ZMS}$. In practice, one should not expect to recover exactly
95\,\%, and a CI for $f_{v,ZMS}$ has to be estimated from the binomial
distribution to account for the limited number of bins\citep{Pernot2022a}.

\subsubsection{Binning/grouping strategies\label{subsec:Binning/grouping-strategies}}

\noindent A sensitive point for the LZMS analysis is the choice of
a binning scheme. The bin size should be small enough to get as close
as possible to individual calibration testing and to provide information
on the localization of any miscalibrated area, but also large enough
to ensure a reasonable power for statistical estimation and testing
of binned statistics. Moreover, stratification of the dataset, if
present, has also to be considered.

\paragraph{Effect of equal-size binning for stratified conditioning variables.}

The equal-size binning scheme is a standard approach implemented for
instance in reliability diagrams\citep{Levi2022}. One has to be aware
that it does not account for the possible stratification of the conditioning
variable. It was shown recently that bin-based statistics in such
conditions are affected by the order of the data in the analyzed dataset.\citep{Pernot2023b_arXiv}
This effect is unavoidable and its impact on the statistics should
be checked, for instance by repeated estimation of the binned statistics
for randomly reordered datasets.

Note that getting a good estimate of $f_{v,ZMS}$ requires contradictory
conditions, i.e. reliable confidence intervals for the binned statistics,
therefore a number of points per bin as large as possible, but also
a number of bins as large as possible. A good balance is obtained
by choosing the number of bins as the square root of the dataset size
($N=M^{1/2})$. The use of this statistic should therefore preferably
be reserved to large datasets, with more than $10^{4}$ points.

\paragraph{Stratified binning.}

For notably stratified conditioning variables, a binning scheme preserving
the strata might be more appropriate than equal-size binning, as it
avoids the splitting of strata into arbitrary bins. However, many
strata might\textcolor{orange}{{} }have sizes too small to enable reliable
statistics. Instead of rejecting these low-counts strata, one can
merge them with their neighbors. I use here an iterative algorithm
where any small stratum (typically less than 100 points) is merged
with the smallest of its neighbors. The value and counts of the strata
are updated according to the relative counts of the merged strata.
This simple merging is iterated until no small stratum is left. The
result is not affected by data ordering. An inconvenience is that
one does not have the control of the number of bins, which might get
too low for a reliable estimation of the $f_{v,ZMS}$ validation statistic.

\paragraph{Choice of conditioning variables or groups for adaptivity.}

\noindent Although using one or several input features as conditioning
variable is the most direct way to test adaptivity, it might not always
be practical, for instance when input features are strings, graphs
or images. In such cases, one might use dimension reduction algorithms
such as t-SNE\citep{Vandermaaten2008,Janet2019} or UMAP\citep{McInnes2018}
in order to define relevant groups. One might also use proxy variables,
latent variables, or even the predicted property value $V$. Using
$V$ answers to the question: \emph{are uncertainties reliable over
the full range of predictions} ? A problem with $V$ is that is is
potentially strongly correlated with $E$ which might lead to spurious
features in the LZMS analysis. For the complementarity of consistency
and adaptivity tests, it is better to use $X$ variables that are
not strongly correlated to $E$ and $u_{E}$. If there is no sensible
way to define a conditioning variable, one might consider adversarial
group validation.

\paragraph{Adversarial groups.\label{par:Adversarial-groups.}}

\noindent One can avoid to choose conditioning variables by designing
random groups to be tested for calibration, in the spirit of \emph{adversarial
group calibration} (AGC). In AGC, the largest calibration error is
estimated over a set of random samples of a given size, for sizes
varying from a small fraction of the dataset to the full dataset.\citep{Chung2020,Zhao2020,Chung2021}
This approach is mostly used to compare datasets, but does not provide
a validation setup. As exposed above, even for fully calibrated datasets,
the amplitude of calibration errors depends also on the group size
and error distribution, which makes comparisons difficult for datasets
with unknown or different distributions. 

It is possible to design an AG \emph{validation} method based of the
$f_{v,ZMS}$ metric. Preliminary test of this approach revealed three
main limitations that make it unpractical: (1) if a dataset presents
localized calibration issues, there is a low probability to randomly
sample groups revealing this problem, and one will get an overly optimistic
diagnostic; (2) random groups are not interpretable, and there is
therefore very few to learn about the origins of miscalibration; and
(3) the computer time for the repeated estimation of converged bootstrap-based
CIs might be prohibitive considering the small information return. 

Another option for validation is to use a conventional AGC curve and
build a probabilistic reference AGC curve, as suggested previously
for confidence curves\citep{Pernot2022c}. In contrast with the latter
case, this probabilistic AGC reference curve is very sensitive to
the choice of a probability distribution for the generated errors,
which might lead to ambiguous diagnostics.

Because of these difficulties, AG validation has not been retained
for the present study. More generally, it has nevertheless to be kept
as an option when the design of adequate conditioning variables is
problematic. Further research to design a robust AGC reference curve
is needed.

\bigskip{}

\noindent \textbf{$\vartriangleright$ Example, continued.} The unbiasedness
and consistency of the QM9 validation set are tested by performing
a LZM/LZMS analysis in $u_{E}$ space with 100 equi-sized bins. The
corresponding $f_{v}$ validation statistics are presented in Fig.\,\ref{fig:fVal}. 

\begin{figure*}[t]
\noindent \begin{centering}
\begin{tabular}{ccc}
\includegraphics[width=0.33\textwidth]{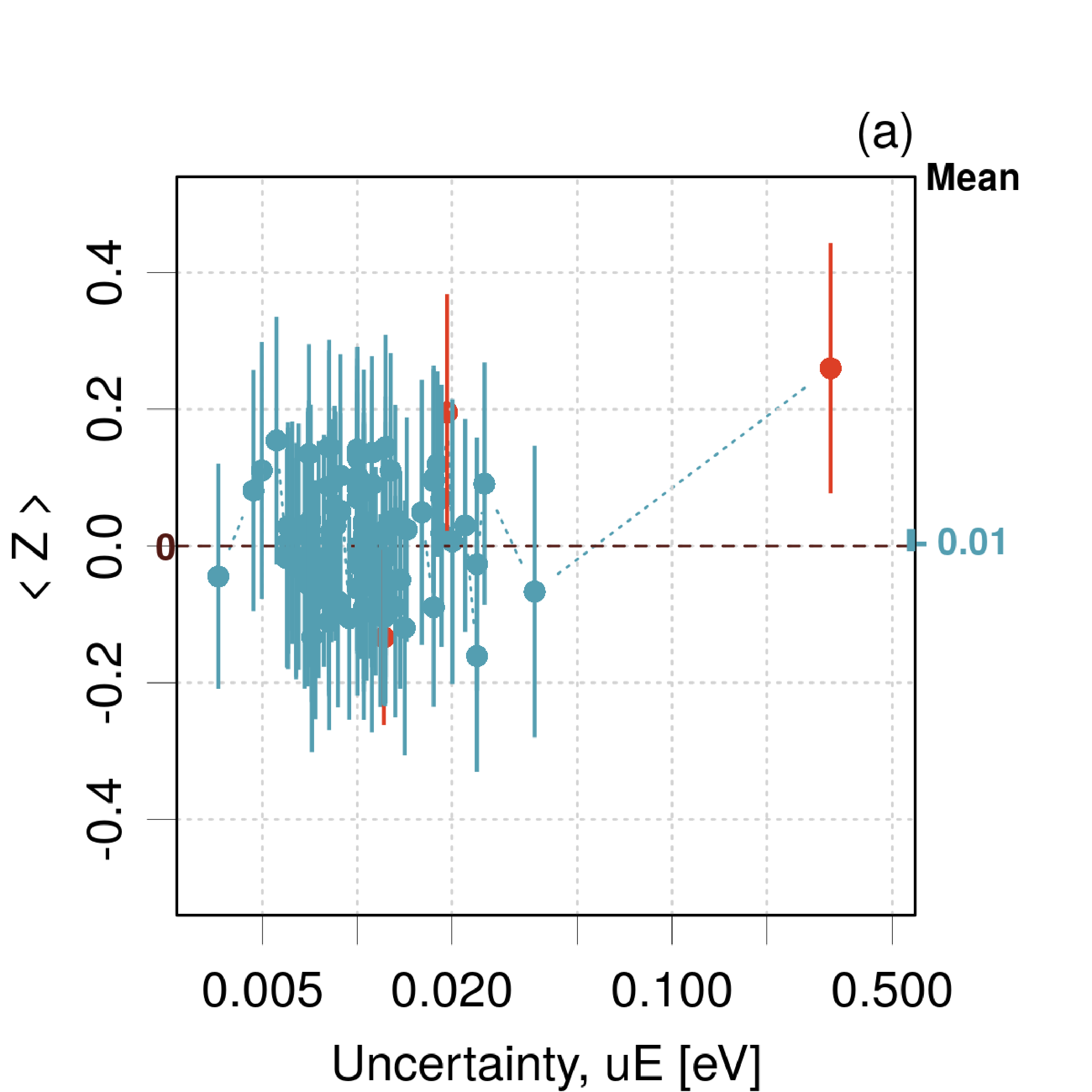} & \includegraphics[width=0.33\textwidth]{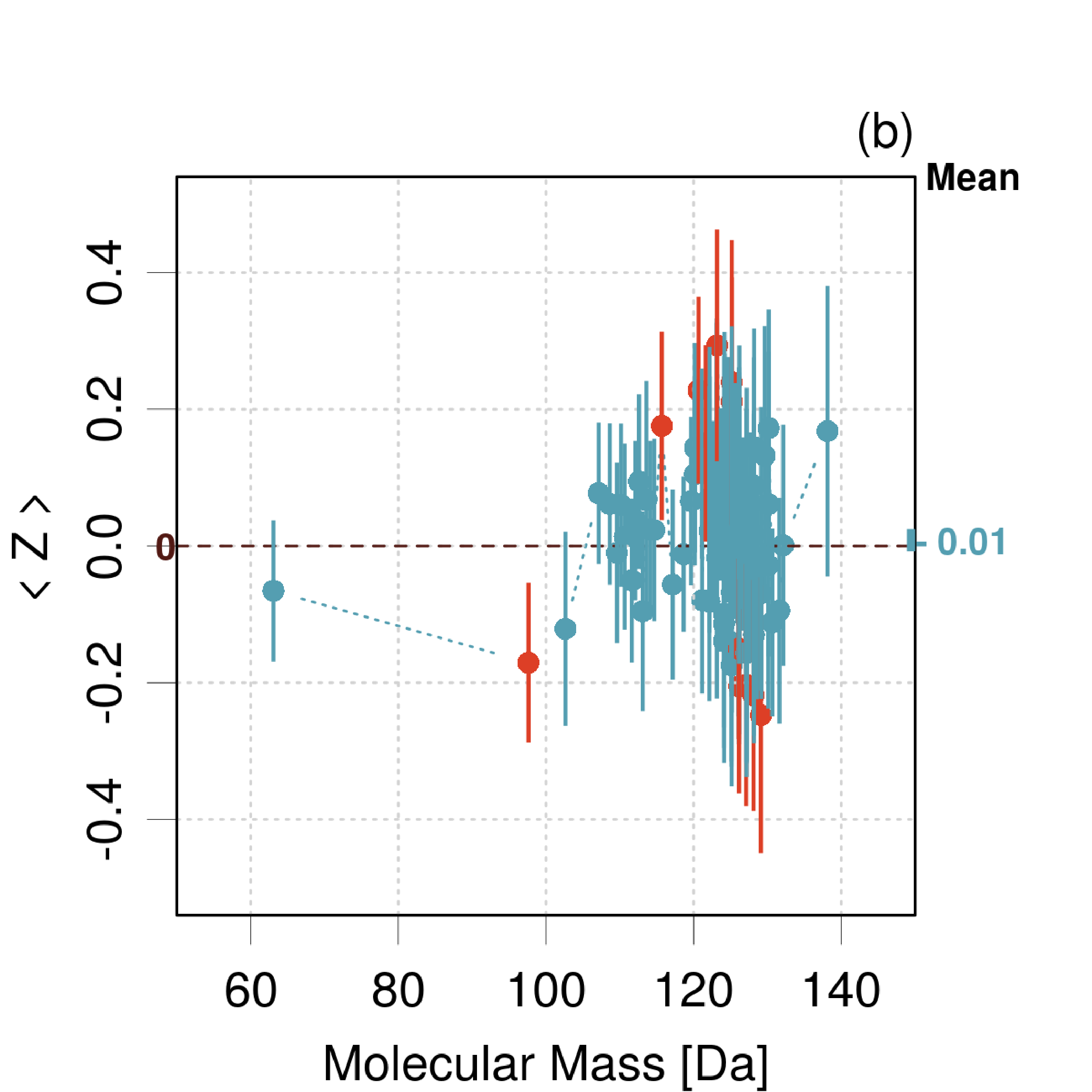} & \includegraphics[width=0.33\textwidth]{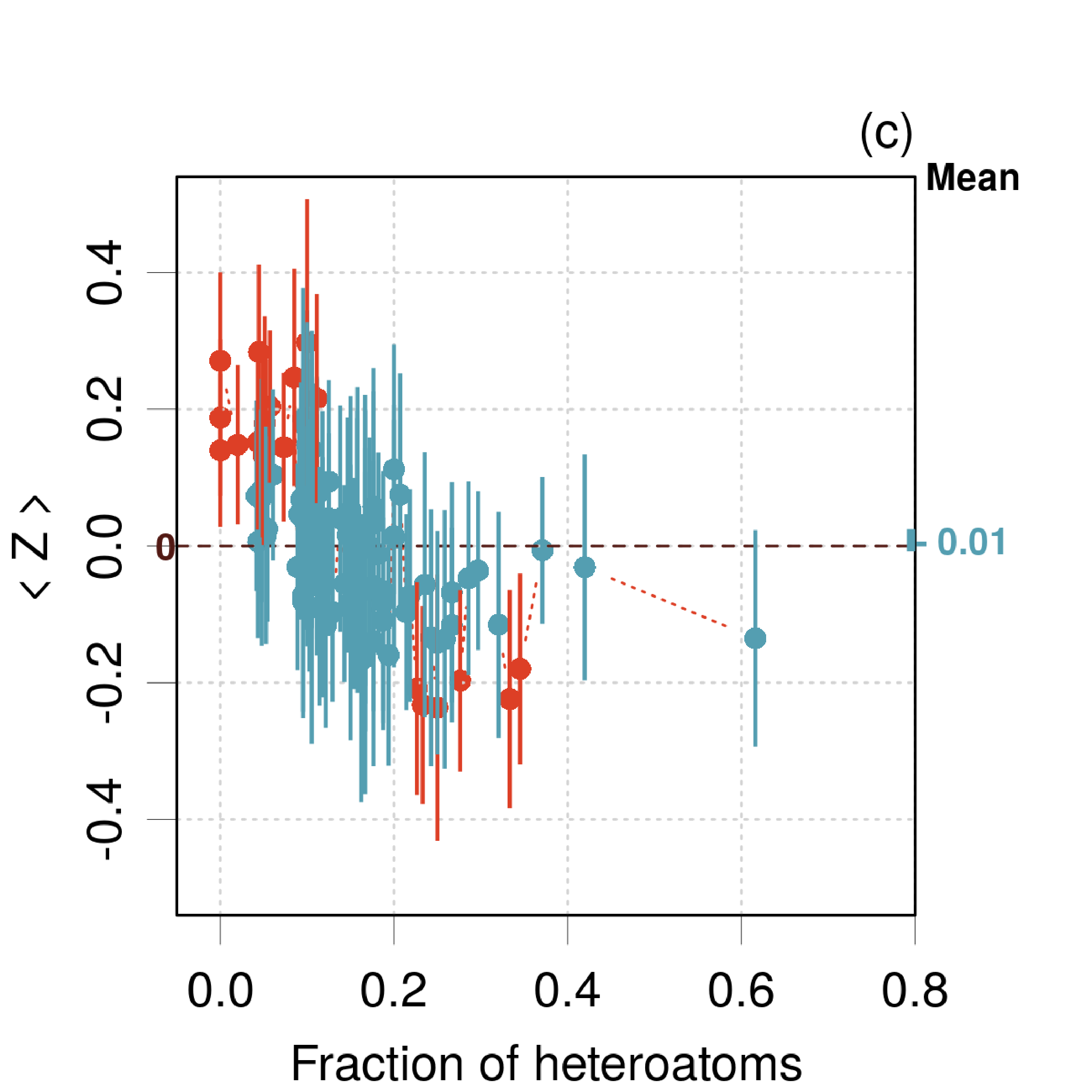}\tabularnewline
\includegraphics[width=0.33\textwidth]{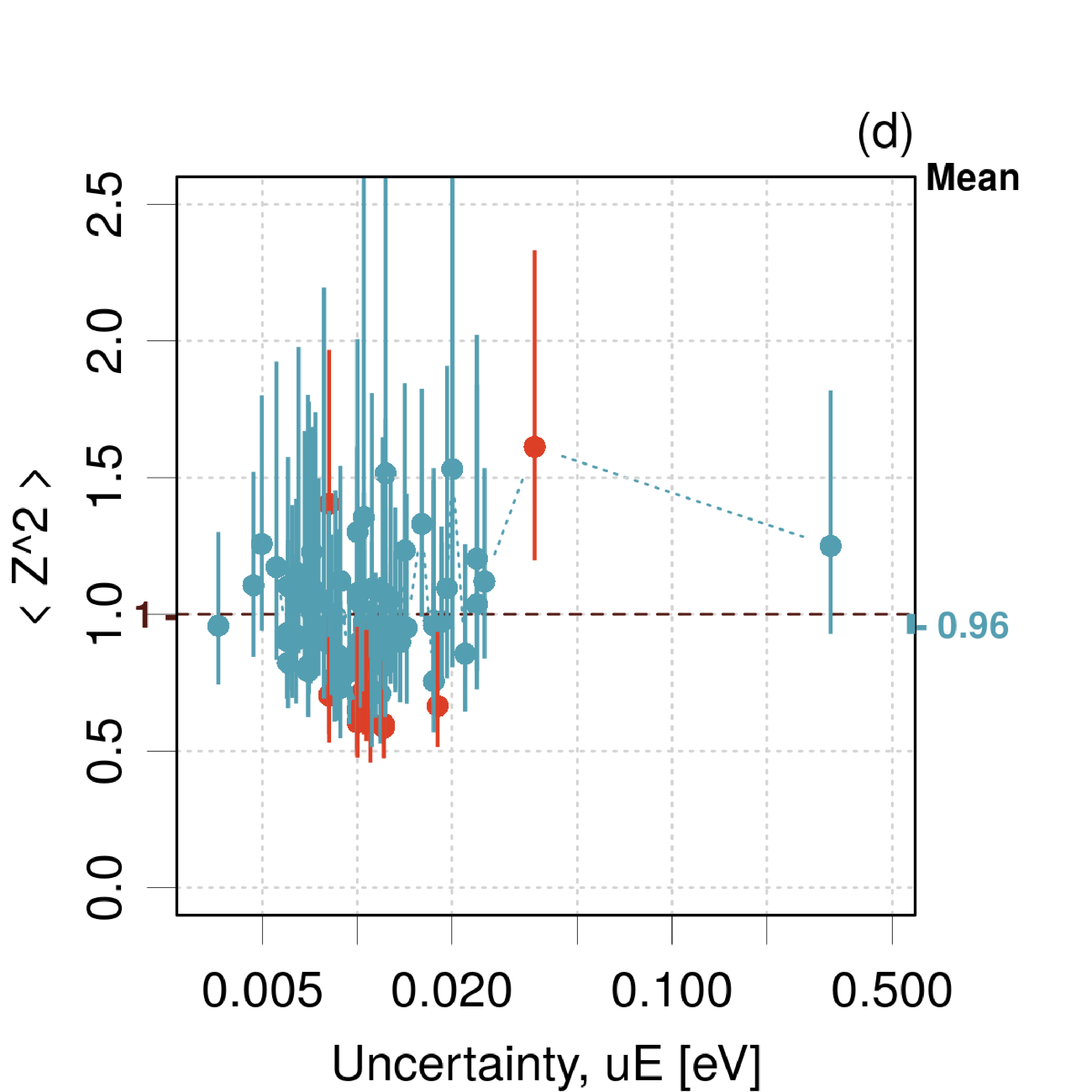} & \includegraphics[width=0.33\textwidth]{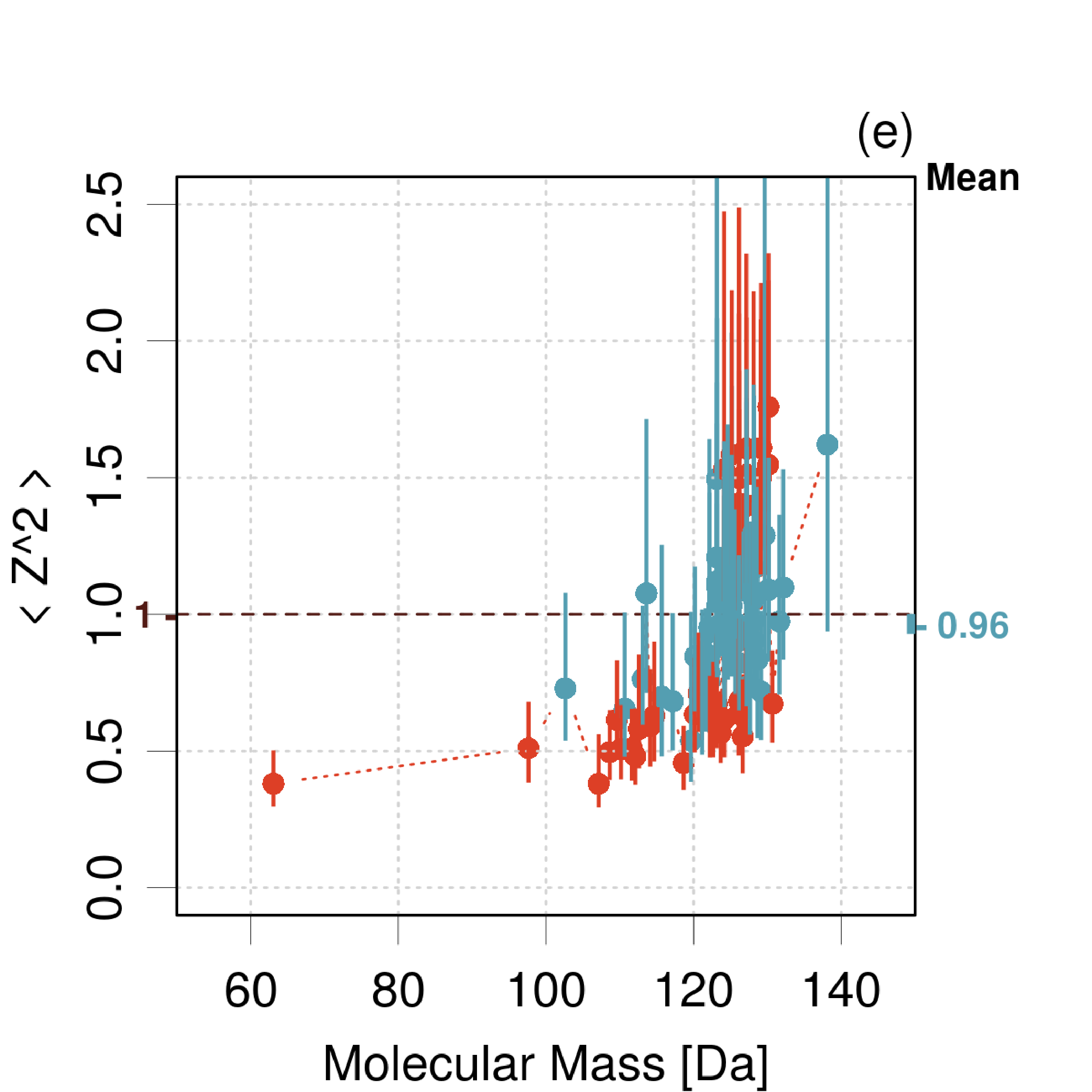} & \includegraphics[width=0.33\textwidth]{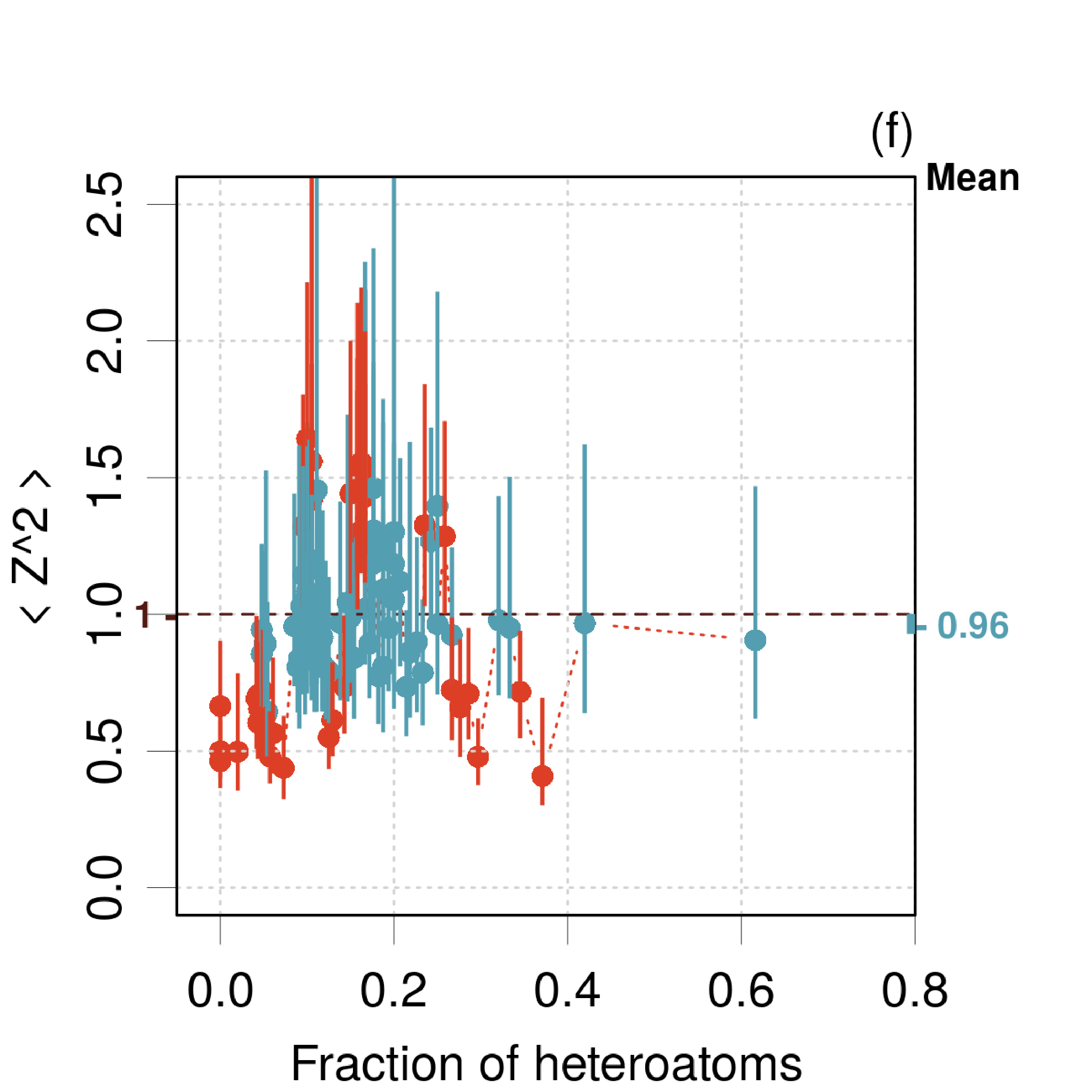}\tabularnewline
\includegraphics[width=0.33\textwidth]{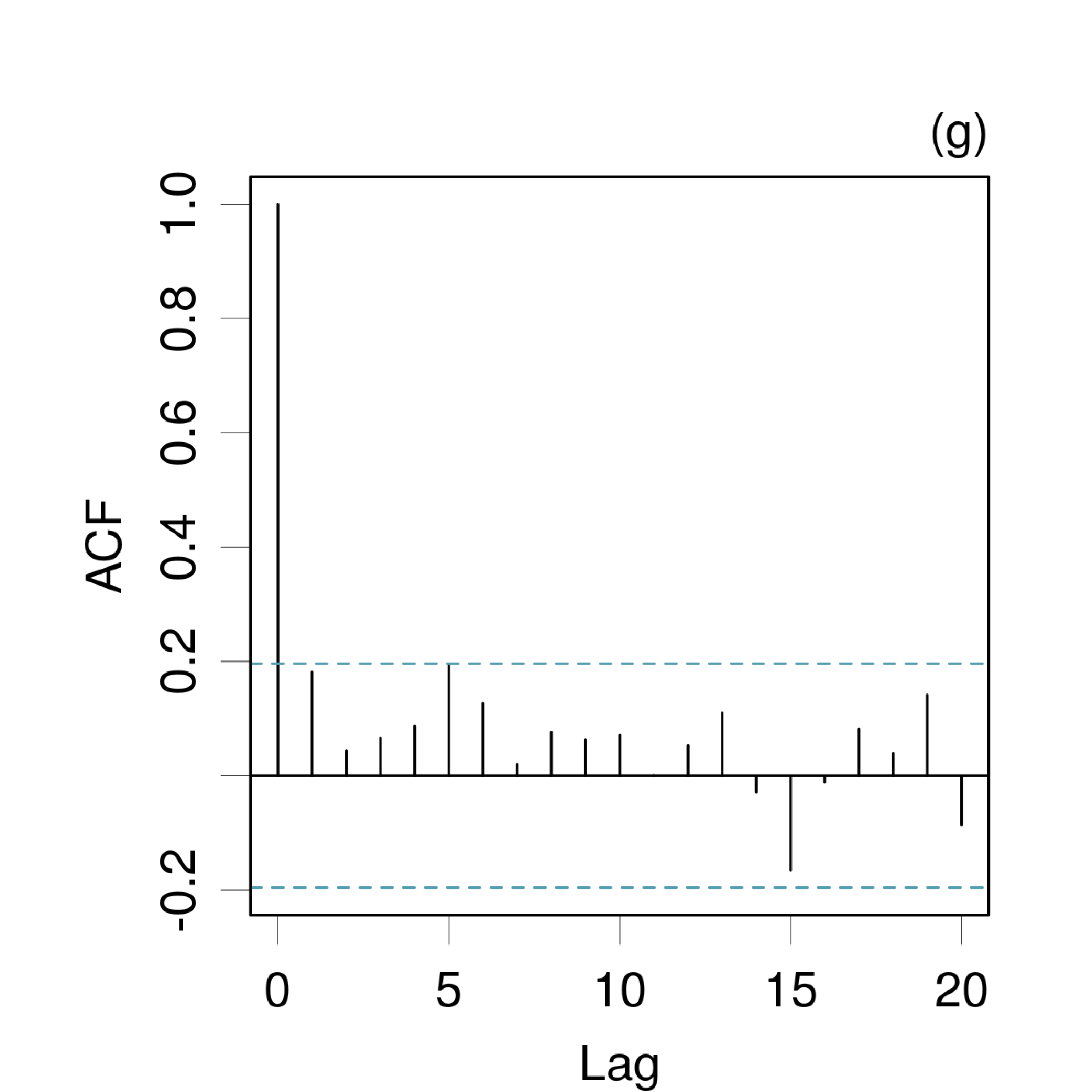} & \includegraphics[width=0.33\textwidth]{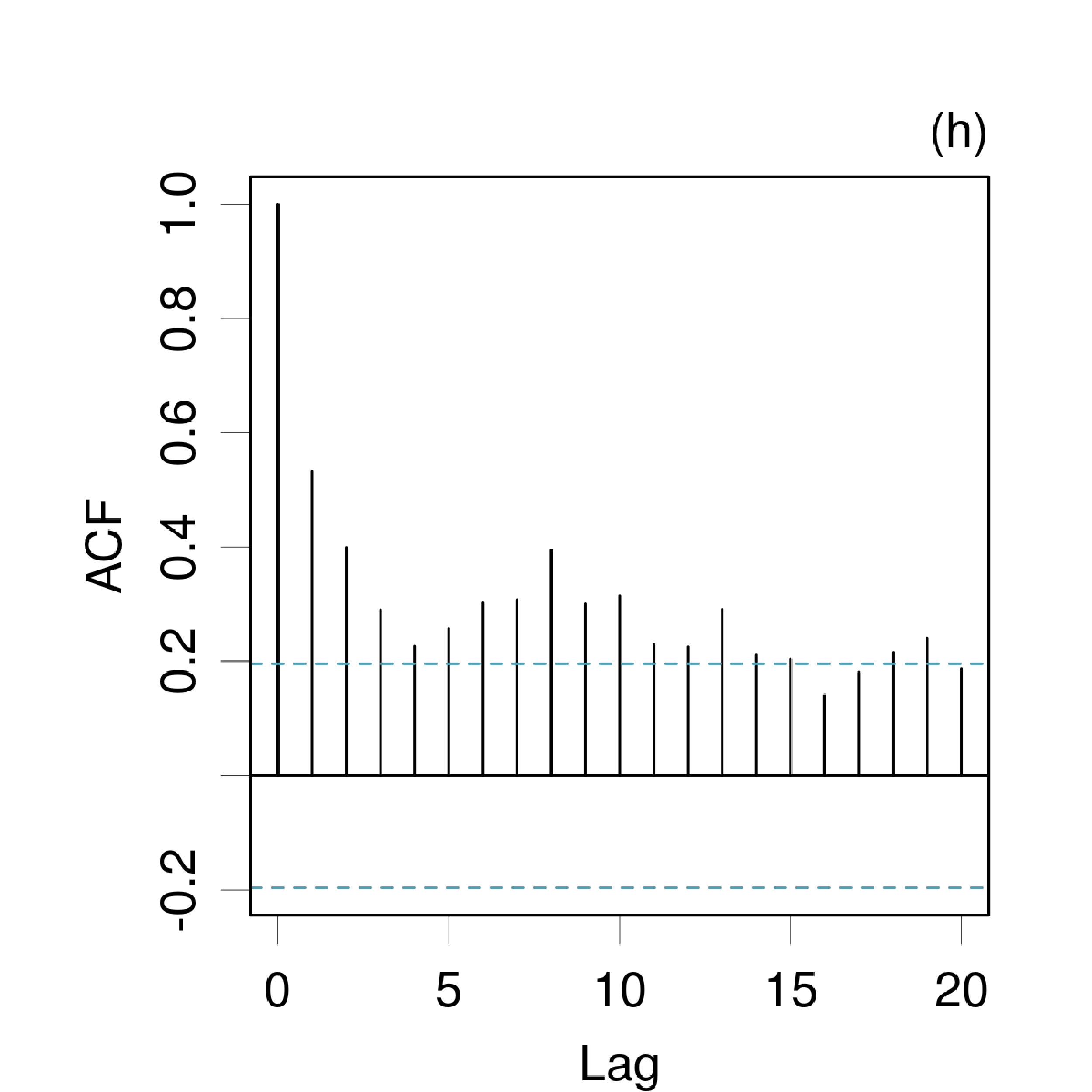} & \includegraphics[width=0.33\textwidth]{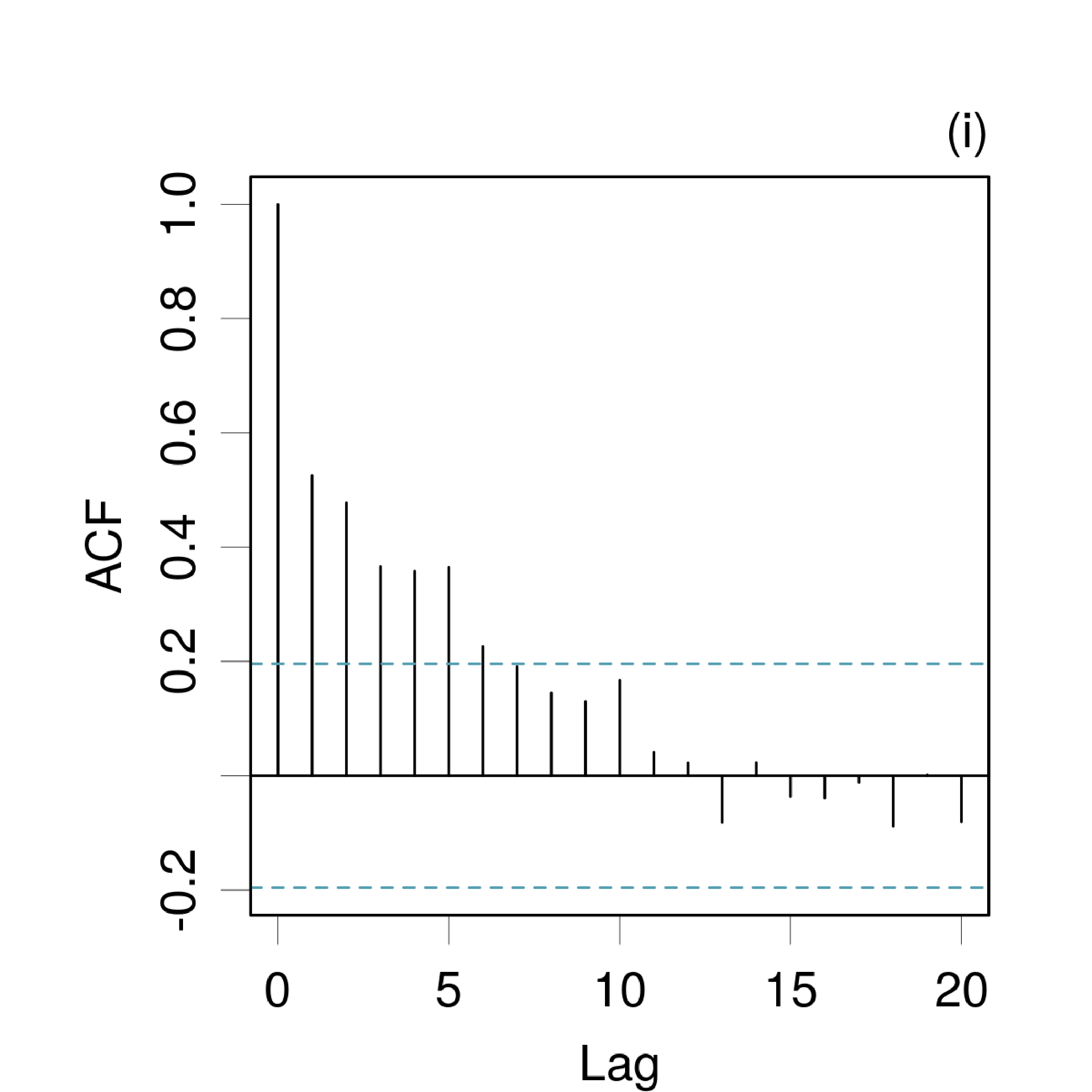}\tabularnewline
\end{tabular}
\par\end{centering}
\caption{\label{fig:eqBin}QM9 validation dataset. Consistency and adaptivity
validation plots based on 100 equal-size bins: LZM and LZMS analyses
and ACF of LZMS vs. $u_{E}$ (a, d, g), molecular mass $X_{1}$ (b,
e, h) and fraction of heteroatoms $X_{2}$ (c, f, i). For the LZM
and LZMS analyses (a-f), the red symbols depict confidence intervals
that do not contain the target statistic (0.0 for $<Z>$; 1.0 for
$<Z^{2}>$), and the mean statistic (for the whole dataset) is reported
in the right margin, with the same color code as for the local statistics.
The corresponding $f_{v}$ statistics are reported in Fig.\,\ref{fig:fVal}.}
\end{figure*}

\begin{figure*}[t]
\noindent \begin{centering}
\includegraphics[width=0.99\textwidth]{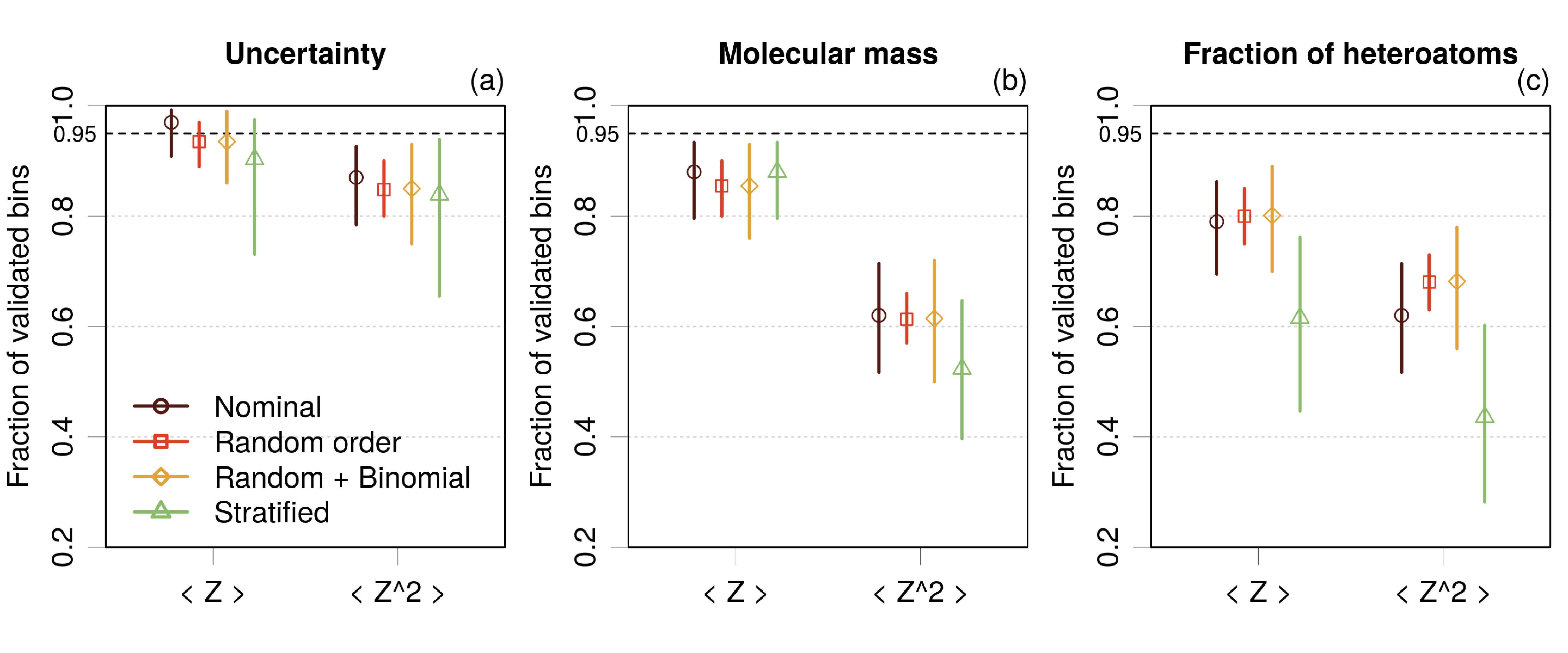}
\par\end{centering}
\caption{\label{fig:fVal}Fraction of validated bins for $<Z>$ (left) and
$<Z^{2}>$ (right) according to three conditioning variables (a-c).
The error bars depict 95\,\% confidence intervals. The fractions
should ideally be compatible with the 0.95 target (horizontal dashed
line). The ``Nominal'' values (black circles) result from equal-sized
binning with 100 bins of the dataset and the error bars are estimated
from a binomial distribution. They summarize the LZM and LZMS analyses
reported in Fig.\,\ref{fig:eqBin}. The ``Random order'' values
(red squares) display the mean and 95\,\% confidence interval for
a random ordering of the dataset (based on 1000 permutations). The
``Random + Binomial'' values (orange diamonds) combine the binomial
uncertainty with the previous values. The ``Stratified'' values
(green triangles) are the statistics for the binning scheme based
on the preservation of strata, with binomial uncertainty. They summarize
the LZM and LZMS analyses reported in Fig.\,\ref{fig:strata}.}
\end{figure*}

One sees for $<Z>$ {[}Fig.\,\ref{fig:eqBin}(a){]} that there are
no outstanding deviations of the binned statistic, except for the
rightmost point (in red), in agreement with the observation on the
previous \emph{z}-scores plots {[}Fig.\,\ref{fig:Plot-of-errors}(a){]}.
The fraction of valid intervals (in blue) is high ($f_{v,ZM}=0.97$)
and in statistical agreement with its target value of 0.95. For $<Z^{2}>$
{[}Fig.\,\ref{fig:eqBin}(d){]}, one observes slightly more deviant
bins, with $f_{v,ZMS}=0.86$, but no major trend, as confirmed by
the ACF analysis {[}Fig.\,\ref{fig:eqBin}(g){]}. Consistency is
therefore not perfect but without strong local miscalibration. As
the uncertainties are stratified, one should evaluate the impact of
data ordering on these statistics. 

For this, $f_{v,ZM}$ and $f_{v,ZMS}$ have been simulated for 1000
random samplings of the data order and summarized by their mean value
and 95\,\% confidence intervals {[}red square in Fig.\,\ref{fig:fVal}(a){]}.
To account for the finite number of bins, each value has also been
perturbed by binomial noise {[}orange diamonds in Fig.\,\ref{fig:fVal}(a){]}.
In the present case, the dispersion due to data reordering is small
when compared to the binomial uncertainty for both statistics. The
data ordering uncertainty is not sufficient to get an overlap of the
confidence interval for $f_{\nu,ZMS}$ with the target coverage (0.95),
validating the conclusions of the nominal LZMS analysis. 

Adaptivity is tested using the same protocol. For the molecular mass
{[}Fig.\,\ref{fig:eqBin}(b,e,g){]}, the fraction of biased bins
reaches 12\,\% ($f_{v,ZM}=0.88$) and the fraction of deviant bins
for $<Z^{2}>$ is about 40\,\% ($f_{v,ZMS}=0.6$). There is a strong
predominance of deviant bins for the masses below 120 Da, where the
small values of the statistic point to overestimated uncertainties.
In consequence, the ACF of the LZMS series presents a slow decay,
to be compared with the one obtained for $u_{E}$. If one accepts
that there is no strong bias of $Z$ in this area, values of $<Z^{2}>$
around 0.5 can be interpreted as an excess factor of $1/\sqrt{0.5}\simeq1.4$
for the uncertainties. Here again, despite the notable stratification
of the molecular masses, the ordering of the data has not a strong
impact on the $f_{v}$ statistics {[}Fig.\,\ref{fig:fVal}(b){]}. 

For the fraction of heteroatoms {[}Fig.\,\ref{fig:eqBin}(c,f,i){]},
the LZM analysis displays the same trend from positive to negative
$<Z>$ values as observed in Fig.\,\ref{fig:Plot-of-errors}(c),
with a sub-optimal fraction of valid bins ($f_{v,ZM}=0.80$). The
LZMS analysis reveals clusters of deviant bins at several spots along
the $X_{2}$ axis, which is reflected in a slowly decreasing ACF.
The fraction of valid bins is small ($f_{v,ZMS}=0.62$). Here again,
despite the strong stratification of $X_{2}$, the $f_{v,ZMS}$ statistic
is not strongly affected by the reordering perturbation {[}Fig.\,\ref{fig:fVal}(c){]}. 

\begin{figure*}[t]
\noindent \begin{centering}
\begin{tabular}{ccc}
\includegraphics[width=0.33\textwidth]{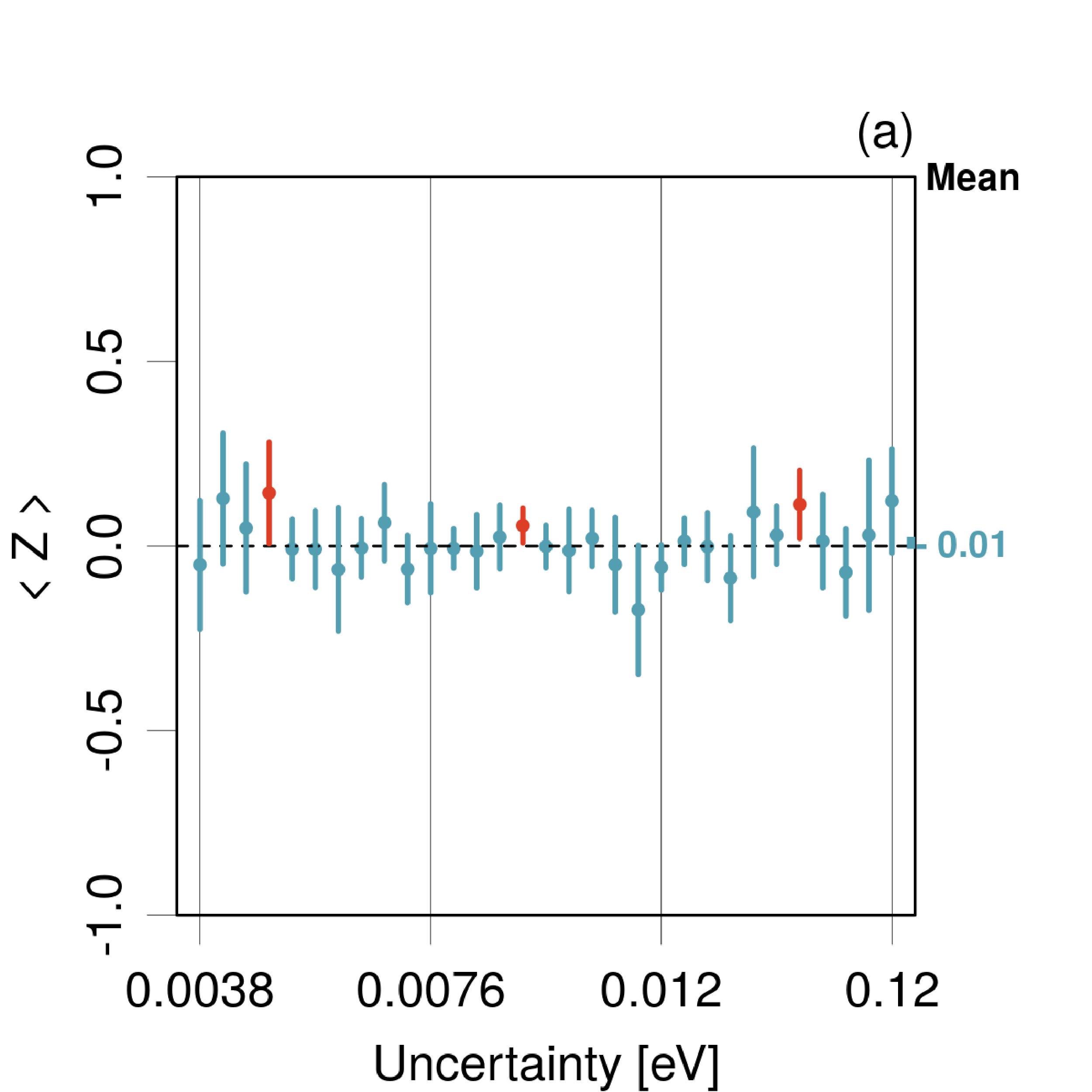} & \includegraphics[width=0.33\textwidth]{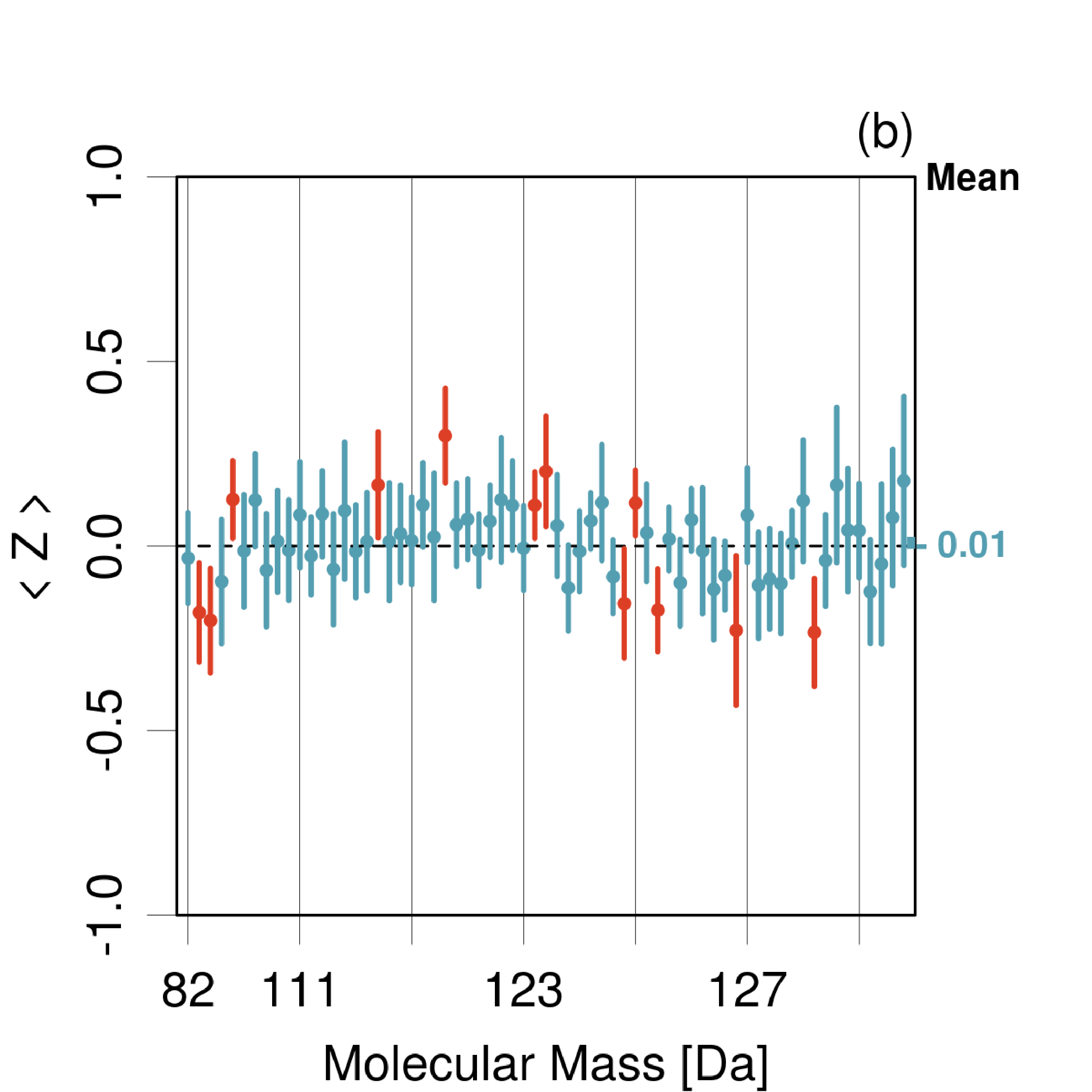} & \includegraphics[width=0.33\textwidth]{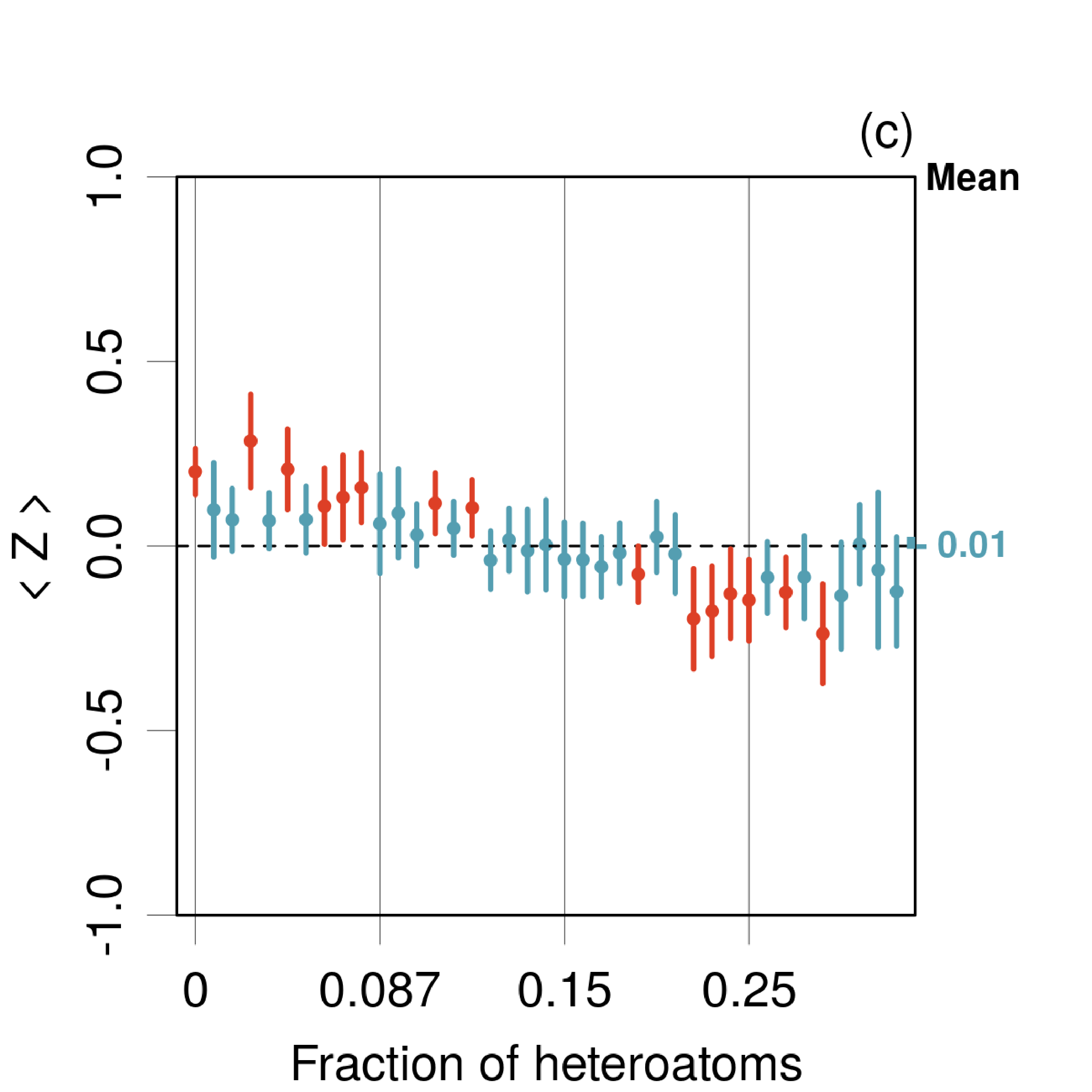}\tabularnewline
\includegraphics[width=0.33\textwidth]{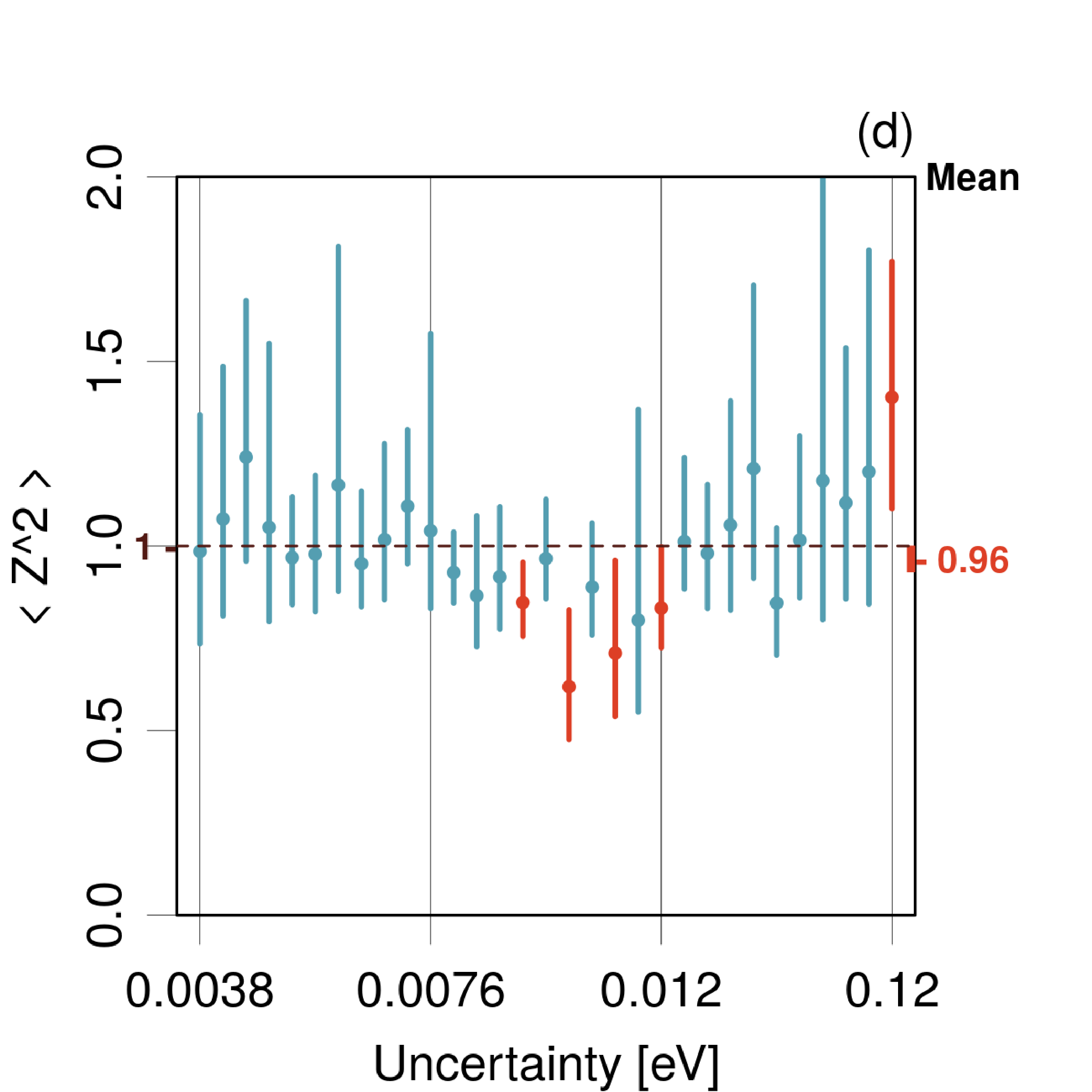} & \includegraphics[width=0.33\textwidth]{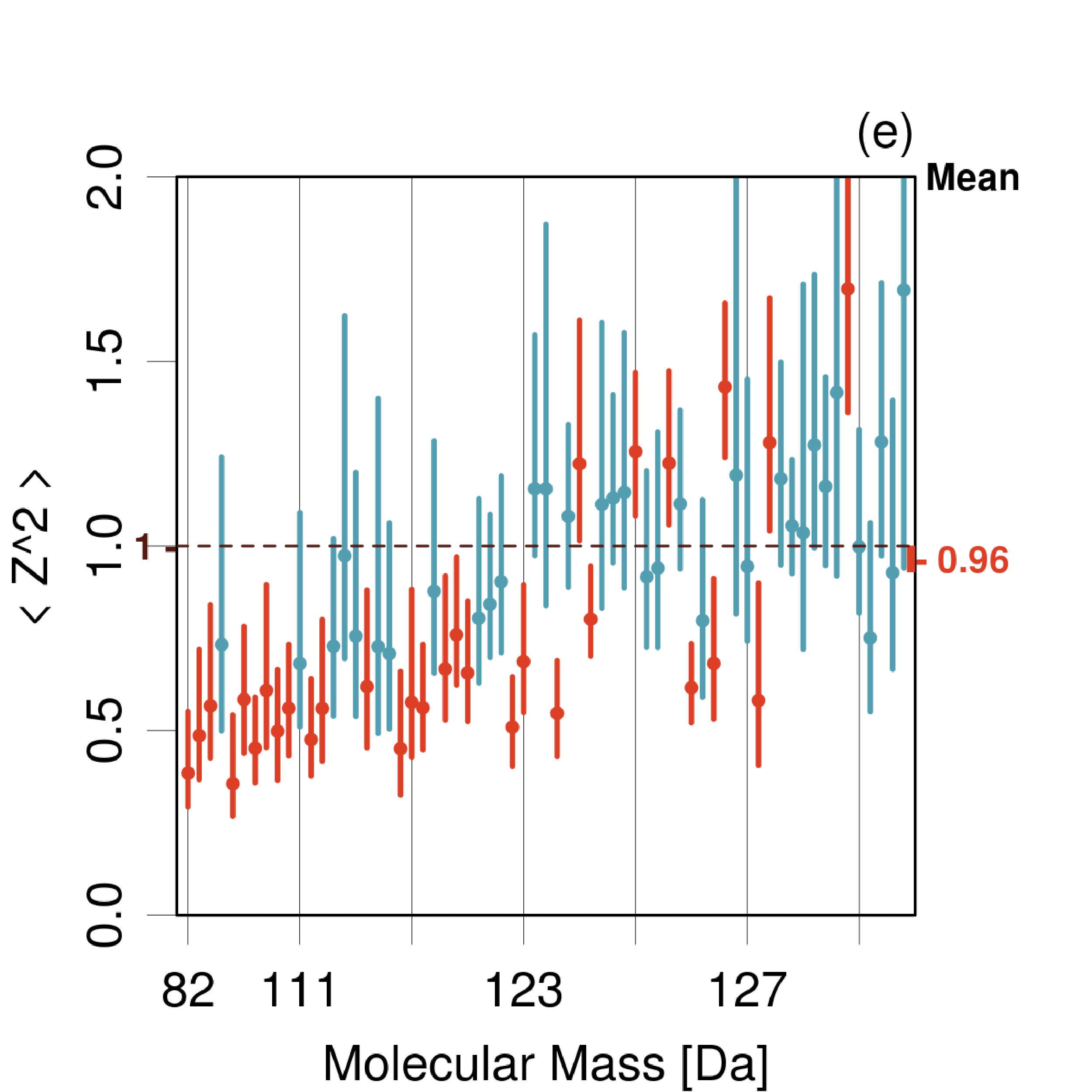} & \includegraphics[width=0.33\textwidth]{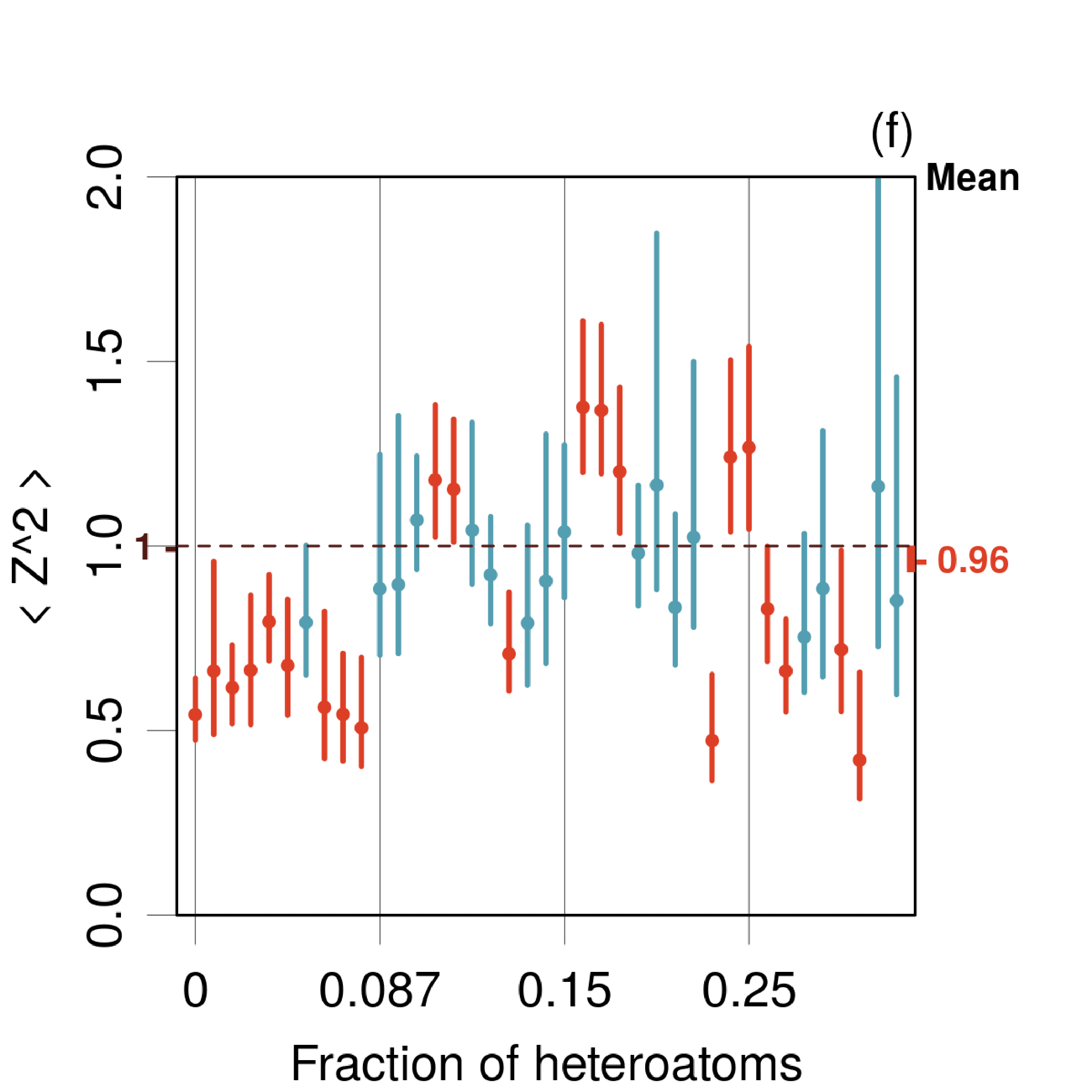}\tabularnewline
\end{tabular}
\par\end{centering}
\caption{\label{fig:strata}QM9 validation dataset. LZM, LZMS analyses vs.
$u_{E}$ (a, d), molecular mass (b, e) and fraction of heteroatoms
(c, f). The data have been aggregated to get a minimum of 100 points
per stratum. The red symbols depict confidence intervals that do not
contain the target statistic (0.0 for $<Z>$; 1.0 for $<Z^{2}>$).
The mean statistic (over the whole dataset) is reported in the right
margin, with the same color code as for the local statistics. The
corresponding $f_{v}$ statistics are reported in Fig.\,\ref{fig:fVal}.}
\end{figure*}

The LZM/LZMS analysis based on stratified binning with a minimum of
100 points per bin is presented in Fig.\,\ref{fig:strata}, and the
$f_{v}$ values are also reported in Fig.\,\ref{fig:fVal}. This
representation is less crowded than the equal-size binning and provides
essentially the same conclusions. One notes more severe values of
the $f_{v}$ statistic for the adaptivity analysis, with larger error
bars due to a smaller number of bins.

All diagnostics based on $E$ and $u_{E}$ conclude therefore to a
good calibration and an acceptable consistency. The main feature revealed
by this analysis is the lack of adaptivity seen by the LZMS analysis
for both molecular mass and fraction of heteroatoms. A major trend
is a significant underestimation of the quality of predictions for
the lighter molecules in the QM9 dataset (below 120 Da) and also for
those with a small fraction of heteroatoms (below 0.1).\hfill{}{
$\boldsymbol{\square}$} 

\subsubsection{Alternative approach: the Local Relative Calibration Error\label{subsec:LRCE}}

\noindent Deriving from the logic behind reliability diagrams, a popular
measure to assess the error in calibration is the Expected Normalized
Calibration Error (ENCE)\citep{Levi2022}, which averages the absolute
relative calibration errors over the bins
\begin{equation}
ENCE=\frac{1}{N}\sum_{i=1}^{N}|RCE_{i}|\label{eq:ENCE}
\end{equation}
where $N$ is the number of bins, $RCE_{i}$ is the Relative Calibration
Error in bin $i$
\begin{equation}
RCE_{i}=\frac{RMV_{i}-RMSE_{i}}{RMV_{i}}
\end{equation}
$RMSE_{i}$ is the root mean squared error for bin $i$, and $RMV_{i}$
is the root mean variance ($u_{E}^{2}$) in bin $i$. 

In the usual applications of ENCE, the bins are based on uncertainty,
so that the ENCE is a measure of consistency. However, Eq.\,\ref{eq:ENCE}
is valid for any binning scheme, and therefore the ENCE can also be
used as an adaptivity measure. 

In this context, the binned RCE offers an alternative to the \emph{z}-scores
formulation and can be used to establish conditional equations similar
to Eqns.\,\ref{eq:LZMS-cons}-\ref{eq:LZMS-adapt} 
\begin{align}
(RCE|u_{E}=\sigma) & \simeq0,\,\forall\sigma>0\label{eq:LRCE-cons}
\end{align}
and
\begin{align}
(RCE|X=x) & \simeq0,\,\forall x\in\mathcal{X}\label{eq:LRCE-adap}
\end{align}
This defines the Local RCE (LRCE) analysis that can be implemented
through data binning according to any conditioning variable, as for
the LZMS analysis. 

This formulation could be more appealing to users familiar with the
ENCE, despite the underlying problem mentioned for average calibration
that the \emph{RMSE} and \emph{RMV} values are insensitive to the
pairing of errors and uncertainties. The ZMS approach is therefore
more robust. However, the most important goal at the present stage
of ML-UQ development is for practitioners to assess adaptivity, be
it by LRCE or LZMS. Besides, a LRCE analysis could be more consistent
with existing ENCE-based toolboxes, such as the Uncertainty Toolbox\citep{Chung2021}.

\section{Conclusions\label{sec:Conclusion}}

\noindent The concept of conditional calibration enables to define
two aspects of local calibration: consistency, which assesses the
reliability of UQ metrics across the range of uncertainty values,
and adaptivity, which assesses the reliability of UQ metrics across
the range of input features. As the validation of individual calibration
is practically impossible, one has to rely on validation methods based
on local or group calibration, making consistency and adaptivity complementary
validation targets. 

Consistency and adaptivity can be tested by binned statistics such
as the mean of squared \emph{z-}scores $<Z^{2}>$, leading to the
LZMS analysis. Bins with large deviations from the target value (typically
1) and groups of adjacent bins with similar deviations reveal local
calibration errors. The LZMS analysis enables to test conditional
calibration for any conditioning variable, giving access to both consistency
and adaptivity validation. An alternative formulation based on local
relative calibration errors (LRCE) could also be considered. A validation
metric $f_{v}$ based on the proportion of bins with the confidence
interval of a statistic containing its target value was proposed.
The focus of this study is on variance-based UQ metrics, but this
validation framework can easily be extended to interval-based UQ metrics\citep{Pernot2022a,Pernot2022b}.

These methods were applied to a representative example issued from
a recent study by Busk \emph{et al.}\citep{Busk2022} about atomization
energies from the QM9 dataset, revealing a good average calibration,
a slightly sub-optimal consistency, and a problematic adaptivity,
either in the molecular mass space or the heteroatoms fraction space.
This dataset presents several sources of stratification, and it was
shown that the uncertainty due to the interplay of equal-size binning
with data ordering expected for stratified conditioning variables
is not dominant for the statistics considered here. An alternative
strata-based binning LZMS approach led to similar diagnostics, with
the inconvenience of a larger uncertainty on the validation statistics
due to the smaller numbers of bins. 

Up to now, ML-UQ validation studies in chemical and materials sciences
are mainly focused on consistency. This covers somehow the concerns
of ML-UQ designers who want the reliability of all uncertainties,
either small or large. It was shown however that a positive consistency
diagnostic does not augur of a positive adaptivity diagnostic, and
therefore that a good consistency does not imply a good individual
calibration. There is therefore a strong need that adaptivity be also
systematically considered in ML-UQ studies, notably for final users,
who expect the reliability of uncertainty for individual predictions,
throughout the input features space. 

\section*{Acknowledgments}

\noindent I warmly thank J. Busk for providing me the data for the
running example of this study.

\section*{Author Declarations}

\subsection*{Conflict of Interest}

The author has no conflicts to disclose.

\section*{Code and data availability\label{sec:Code-and-data}}

\noindent The code and data to reproduce the results of this article
are available at \url{https://github.com/ppernot/2023_Adaptivity/releases/tag/v1.1}
and at Zenodo\citep{Adaptivity}. The
\texttt{R},\citep{RTeam2019} \href{https://github.com/ppernot/ErrViewLib}{ErrViewLib}
package implements the validation functions used in the present study,
under version \texttt{ErrViewLib-v1.7.3} (\url{https://github.com/ppernot/ErrViewLib/releases/tag/v1.7.3}),
also available at Zenodo\citep{ErrViewLib}.

\bibliographystyle{unsrturlPP}
\bibliography{NN}

\end{document}